\documentclass{article}

\usepackage[english]{babel}

\usepackage[letterpaper,top=2cm,bottom=2cm,left=3cm,right=3cm,marginparwidth=1.75cm]{geometry}
\usepackage{booktabs}
\usepackage{floatrow}
\usepackage{listings}
\usepackage{amsmath}
\usepackage{graphicx}
\usepackage{subcaption}
\usepackage{multirow}
\usepackage{amsmath}  
\usepackage{amsfonts}  
\usepackage[colorlinks=true, allcolors=blue]{hyperref}

\title{Capturing Momentum: Tennis Match Analysis
	Using Machine Learning and Time Series Theory}
\author{Jingdi Lei, Tianqi Kang, Yuluan Cao,Shiwei Ren}

\begin{document}
\maketitle


\begin{abstract}
	This paper presents an analysis on the momentum of tennis match. And due to it's Generalization performance, it can be helpful in constructing a system to predict the result of sports game and analyze the performance of player based on the Technical statistics. We First use hidden markov models to predict the momentum which is defined as the performance of players. Then we use Xgboost to prove the significance of momentum. Finally we use LightGBM to evaluate the performance of our model and use SHAP feature importance ranking and weight analysis to find the key points that affect the performance of players.
	
\end{abstract}


\maketitle  

\section{Introduction}
\subsection{Background and Problem Statement}
\begin{figure}[htbp]
	\centering
	\includegraphics[width=.5\textwidth]{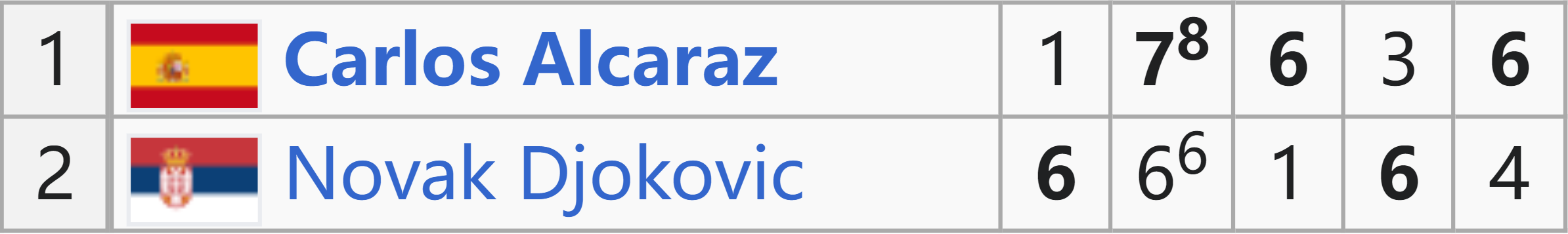}
	\caption{2023 Wimbledon Gentlemen's final scores}\label{fig:2023match}
\end{figure}

In the 2023 Wimbledon Gentlemen's final, Spanish starlet Alcaraz defeated Djokovic. As shown in Figure \ref{fig:2023match}, the swings in the final scores not only made the spectators excited and nervous, but also reflected the multiple shifts of momentum during the match. This begged the question of whether momentum could play a role in a match. Studying this question can better predict game flow and scores, as well as help coaches and players with post-match review and targeted training.

Considering the background, we are supposed to solve the following problems:
\begin{itemize}
    \item Build a model to captures the flow of play as match points occur and visualize it. Quantitatively describe the performance of the player. The server should be considered in the model.
    \item Use the model to prove the significance of momentum. 
    \item Explore some indicators for determining when momentum changes, that is, the flow of play changes from favoring one player to the other. Use the data to build a volatility prediction model to predict these "momentum" swings, and identify the most relevant factors. Based on the models, give players suggestions on how to prepare for a new match against a different opponent, based on the "momentum" swings in previous games.
    \item Test the accuracy, universality, generalization of the model on other matches. If sometimes it is not good, what potential features can be added in future models?
\end{itemize}
\subsection{Related Work}

~~~~Many scholars have begun to conduct tennis-related research very early. Their research on tennis involves analysis of players' technical and tactical levels, influencing factors of tennis matches, prediction of tennis results, and tennis betting. These papers mainly focus on the prediction of tennis results. The most common one is the prediction model represented by the \textbf{Markov model}.

Barnett, T. and Brown, A. (2006) introduced the use of \textbf{Markov chain} to calculate the winning probability of the game, set, and match based on the probability of each player's serve and score. They assumed that the probability of scoring a serve is a constant in their calculations\cite{1}.
Newton (2006) used a \textbf{Markov model} to predict the outcome of a match in which the probability of a player's serve scoring was updated every round. The probability of defeating an opponent and winning the championship were also calculated at the end of each round \cite{2}.
Based on this research \cite{2}, Newton (2009) corrected the probability of A's serve scoring in an A vs. B match by taking into account the opponent's ability. The corrected probabilities were then used in the previous Markov chain \cite{3}.

Barnett and Clarke considered predicting the outcome of a tennis match using a modified Markov chain, where he considered the probability of a player's serve scoring a point is not a constant. The paper assumes that if player A leads, the probability of A winning a set increases by $\alpha$, and the probability of the set score is adjusted according to the set score\cite{4}.

For \textbf{modeling momentum}, many studies investigated momentum shifts. Richardson et al. \cite{5}, Jones and Harwood  \cite{6} reported there were momentum shifts by interviewing players. 
Paper \cite{7} introduced Markov chain processes for modeling unobserved sequence-dependent processes, often called \textbf{Hidden Markov Models (HMMs)}.
The article introduced the basic theory and concepts of Hidden Markov Model and its wide range of applications in practical problems.
Ötting et al. \cite{8} analyzed momentum changes in football matches. By analyzing minute-by-minute summary statistics, this paper investigate the potential occurrence of momentum transfer using Hidden Markov Model (HMM).

\subsection{Our Work}
\begin{figure}[htbp]
	\centering
	\includegraphics[width=\textwidth]{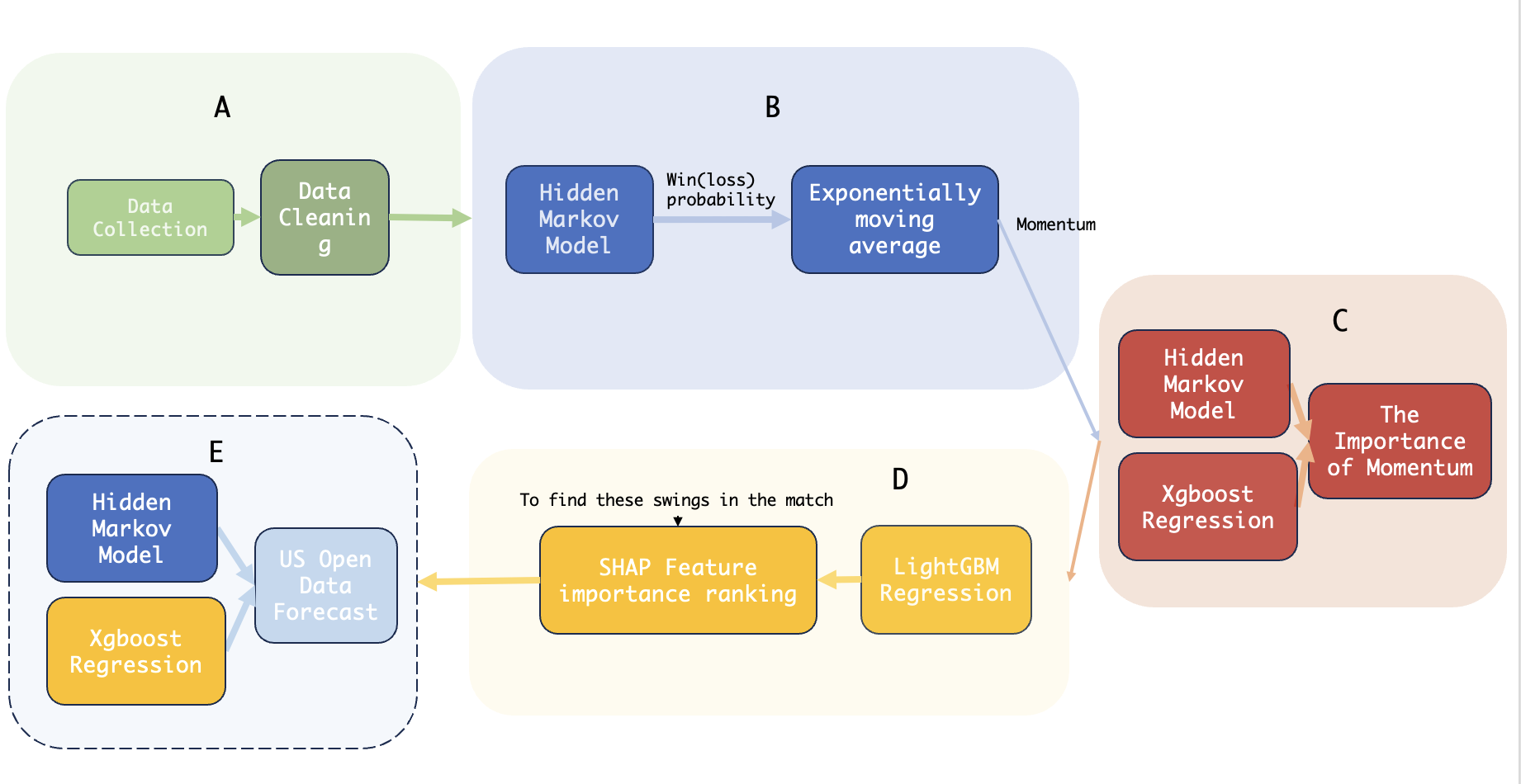}
	\caption{Our work and models in the whole process. A is data cleaning . B presents the method to find the momentum. C shows the method to prove the significance of momentum. D find the importance feature in momentum change. E evaluate the performance of our model}\label{fig:work}
\end{figure}

\section{Preparation for the Models}

\subsection{Data Preprocessing}
First, before building the model, we preprocess the data. Data preprocessing is an indispensable step in the process of data analysis. By cleaning, converting and organizing data, we can improve the quality of the data and ensure the reliability and effectiveness of the analysis.
\begin{itemize}
	\item \textbf{Data Sources:} 
	\begin{enumerate}
		\item The data of the first two rounds of Wimbledon \url{https://github.com/JeffSackmann/tennis_slam_pointbypoint}. We will use the entire data set in Model Evaluation (section \ref{b}).
		\item The data collected from \url{https://www.usopen.org/}. We selected different men's and women's singles matches for prediction to test the model in Model Evaluation (section \ref{d}). Unlike the Wimbledon, the US Open court material is hard. The Figure \ref{fig:usopen} shows the playing field of the US Open.
		\begin{figure}[htbp]
			\centering
			\begin{subfigure}[b]{.4\textwidth}
					\includegraphics[width=\textwidth]{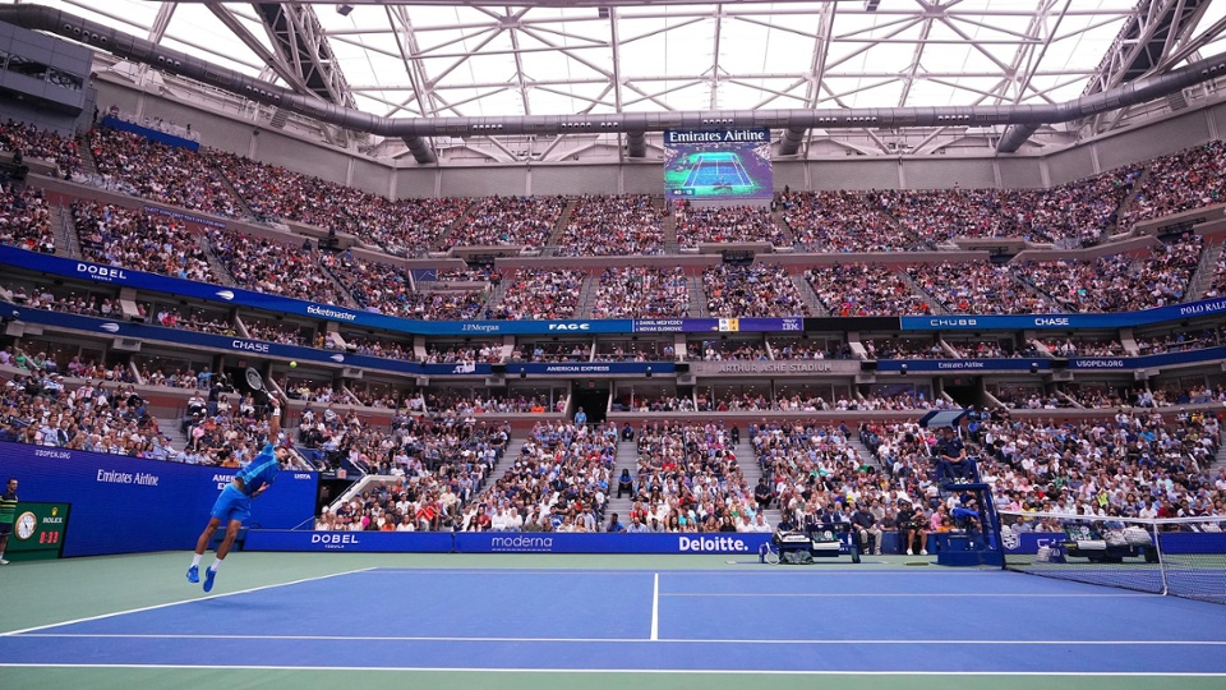}
				\end{subfigure}
			\begin{subfigure}[b]{.4\textwidth}
					\includegraphics[width=\textwidth]{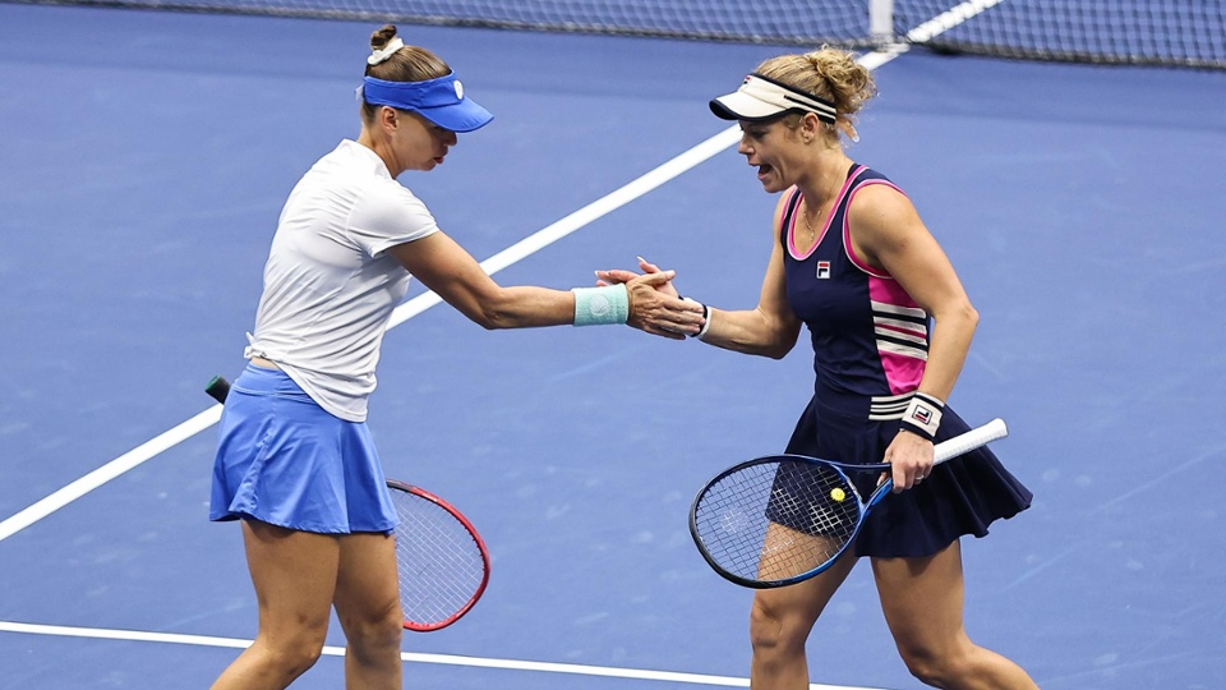}
				\end{subfigure}
			\caption{US Open match map}\label{fig:usopen}
		\end{figure}
	\end{enumerate}
	
\end{itemize}

\begin{figure}[htbp]
	\centering
	\includegraphics[width=1\textwidth]{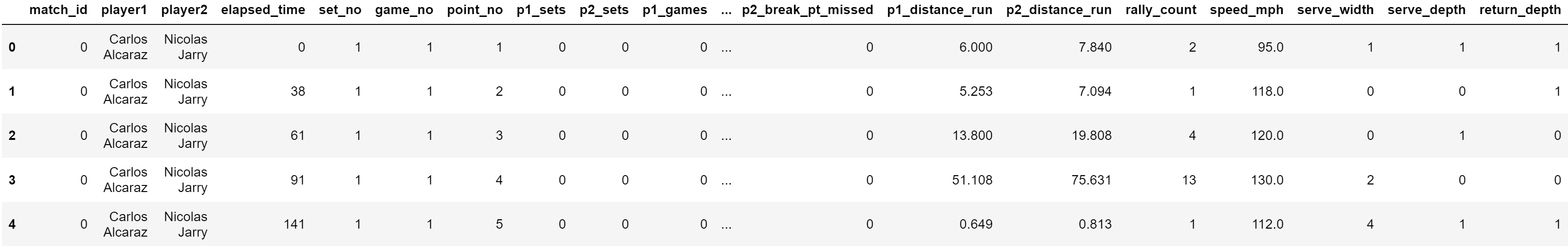}
	\caption{The first four rows of preprocessed data}\label{fig:preprocessed data}
\end{figure}

The preprocessed data is shown in Figure \ref{fig:preprocessed data}.

\section{Momentum Quantification}\label{model}
~~~~First, we did the accumulation of time and calculated some indicators such as the player's scoring rate and break rate so far in the match. Then we combined these indicators with the characteristics of the match at the moment in the data. From four common Markov models, We choose to use \textbf{HMM(Hidden Markov model)} to find the transition probability of the hidden variable, that is the probability of winning or losing. Finally from that probability, we use \textbf{EMA (exponential moving average)} to get momentum.

\subsection{Model Selection}

\subsubsection{Hidden Markov model}
Since state transitions (that is momentum in this problem) cannot be directly observed, we use Hidden Markov Models (abbreviated as HMM). In 1966, LEONARD E. Baum and J. A. EAGON first proposed the hidden Markov model in the paper \cite{9}. In a normal Markov model, the state is directly visible to the observer. In this way, the transition probability of the state is the whole parameter. Whereas in HMM, we do not know the specific state sequence of the model, but only the probability of state transition, that is, the state transition process of the model is unobservable, but some variables affected by the state are visible.  

\subsection{HMM Construction}
\begin{figure}[htbp]
	\centering
	\includegraphics[width=.6\textwidth]{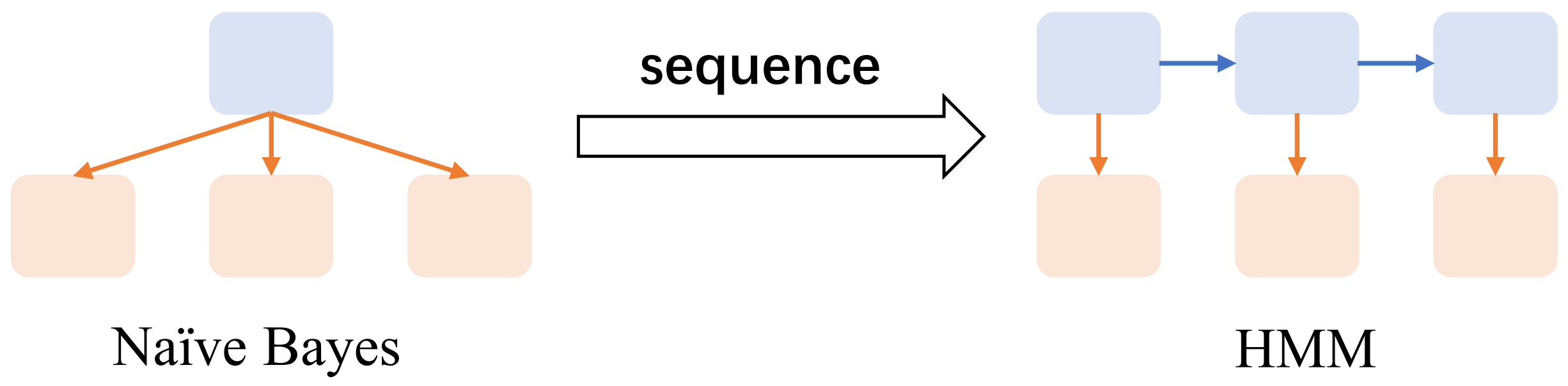}
	\caption{Bayes and HMM}\label{fig:bayes}
\end{figure}

Hidden Markov model is a probabilistic graphical model. We know that machine learning models can be considered from two directions: Frequentist and Bayesian. The core of the Frequentist method is the optimization problem, while in the Bayesian method, the core is the integration problem, and a series of integration methods such as variational inference, MCMC, etc. have also been developed. The most basic model of the probabilistic graphical model can be divided into two aspects: directed graph (Bayesian network) and undirected graph (Markov random field), such as GMM. On these basic models, if there is a correlation between samples , it can be considered that the samples are accompanied by time series information, so the samples are not not independent of each other with full homogeneity of the distribution. This model is called a dynamic model. The hidden variables change over time, so the observed variables also change. The relationship between bayes and HMM is shown in Figure \ref{fig:bayes}.

In HMM, there are three basic assumptions which are already explained in section \ref{Assumptions}:
\begin{enumerate}
	\item Homogeneous Markov assumption
	\begin{equation}\label{Markov}
		\begin{aligned}&
			p(i_{t+1}|i_t,i_{t-1},\cdots,i_1,o_t,o_{t-1},\cdots,o_1)=p(i_{t+1}|i_t)
		\end{aligned}
	\end{equation}
	\item Observational independence hypothesis
	\begin{equation}\label{Markov}
		\begin{aligned}&
			p(o_t|i_t,i_{t-1},\cdots,i_1,o_{t-1},\cdots,o_1)=p(o_t|i_t)
		\end{aligned}
	\end{equation}
	\item Parameter invariance assumption. That is the starting probability distribution $\pi$, the state transition matrix	$A$	and	the emission matrix $B$ remain unchanged.
\end{enumerate}

\begin{figure}[htbp]
	\centering
	\includegraphics[width=.8\textwidth]{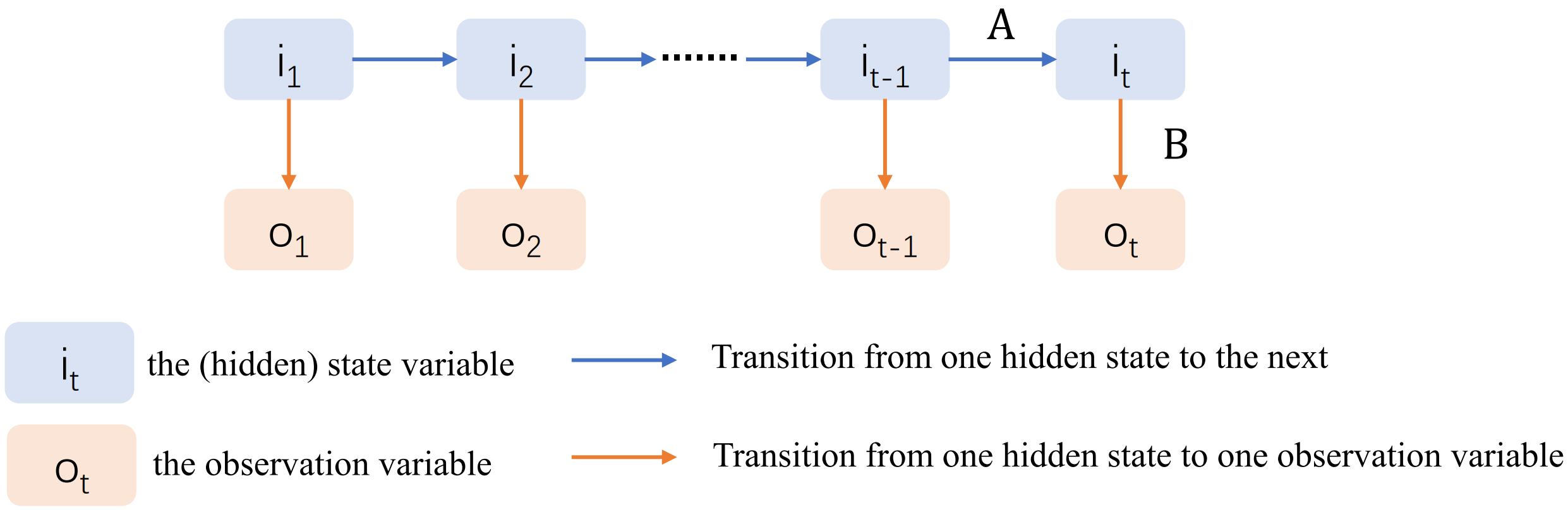}
	\caption{Schematic diagram of the hidden Markov model}\label{fig:HMM}
\end{figure}

Figure \ref{fig:HMM} shows the changes in hidden variables at t moments. 
The hidden Markov model is determined by the initial state vector $\pi$, state transition matrix $A$ and observation probability matrix $B$. $\pi$ and $A$ determine the state sequence, and $B$ determines the observation sequence. Therefore, the hidden Markov model can be represented by a ternary notation, that is $\lambda=(\pi,A,B)$.

We use $o_t$ to denote the observation variable, $O$ the observation sequence, $V={v_1,v_2,\cdots,v_M}$ to denote the value domain of the observation, $i_t$ to denote the state variable, $I$ the state sequence, and $Q={q_1,q_2,\cdots,q_N}$ to denote the value domain of the state variable.

The equations of $\pi$, $A$, $B$ are as follows:
\begin{equation}
	\begin{aligned}&
		\pi=(\pi_{i}) 
		&&&&&&&& \pi_{i}=P(i_{1}=q_{i}),\quad i=1,2,\cdots,N
	\end{aligned}
\end{equation}

\begin{equation}
	\begin{aligned}&
		A=[a_{ij}]_{N\times N}  &&&&&&&&
		a_{ij}=p(i_{t+1}=q_j|i_t=q_i),i=1,2,\cdots,N;j=1,2,\cdots,N
	\end{aligned}
\end{equation}

\begin{equation}\label{B}
	\begin{aligned}&
		B=[b_{k}(k)]_{N\times M} &&&&&&&&
	    b_{j}(k)=p(o_t=v_k|i_t=q_j),k=1,2,\cdots,M;j=1,2,\cdots,N
	\end{aligned}
\end{equation}

$a_{ij}$ represents the probability of being in state $q_i$ a at time t and transitioning to state $q_j$ at time t+1. 
$b_{j}(k)$ represents the probability of generating observation $v_k$ in state $q_j$ at time t (also called generation probability and emission probability).
State $i_t$ is invisible, observation $o_t$ is visible.

Based on the graph structure of the hidden Markov model, the joint probability distribution of all variables is the equation \ref{HMM}.
\begin{equation}\label{HMM}
	\begin{aligned}&
		P(o_1,i_1,\cdots,o_n,i_n)=P(i_1)P(o_1|i_1)\prod_{t=2}^{n}P(i_t|i_{t-1})P(o_t|i_t)
	\end{aligned}
\end{equation}

\subsection{Exponential Moving Average}
EMA (exponential moving average), or exponentially weighted moving average, can be used to estimate the local mean of a variable so that its updates are related to its historical values over a period of time. 

The variable $v$ is recorded as $v_t$ at time $t$, and $\theta_t$ is the value of variable v at time t, that is, $v_t=\theta_t$ when EMA is not used. After using exponential moving average, the update formula of $v_t$ is as follows:
\begin{equation}\label{Markov}
	\begin{aligned}
		v_t=\beta\cdot v_{t-1}+(1-\beta)\cdot\theta_t
	\end{aligned}
\end{equation}
In the above formula, $\beta\in\left[0,1\right)$. $\beta=0$ is equivalent to not using a moving average.

In the optimization algorithm, we generally take $\beta\geq0.9$, and $\begin{aligned}1+\beta+\beta^2+\ldots+\beta^{t-1}=\frac{1-\beta^t}{1-\beta}\end{aligned}$. When t is large enough, $\beta^t\approx0$, it is an exponentially weighted moving average in the strict sense. And in our model, we take $\beta\geq0.5$ for better performance.

\subsection{The Match Flow}
\begin{figure}[htbp]
	\centering
	\begin{subfigure}[t]{.49\textwidth}
                \centering
			\includegraphics[width=\textwidth]{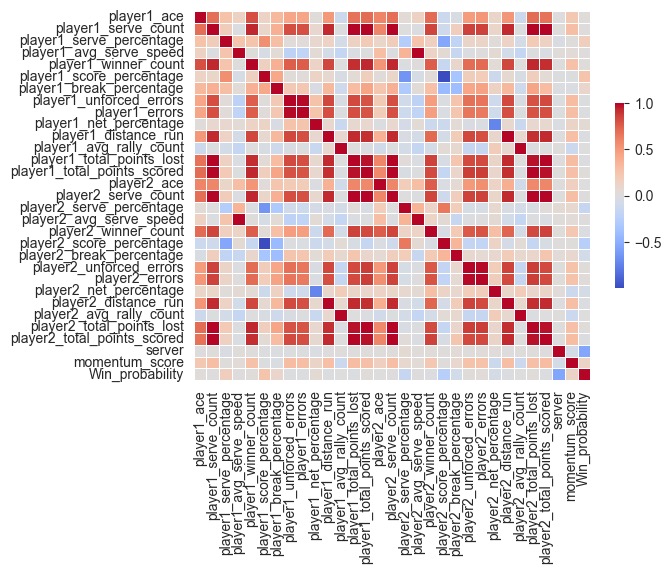}
			\caption{Pearson correlation coefficient}\label{fig:Pearson}
		\end{subfigure}
	\begin{subfigure}[t]{.49\textwidth}
                \centering
			\includegraphics[width=\textwidth]{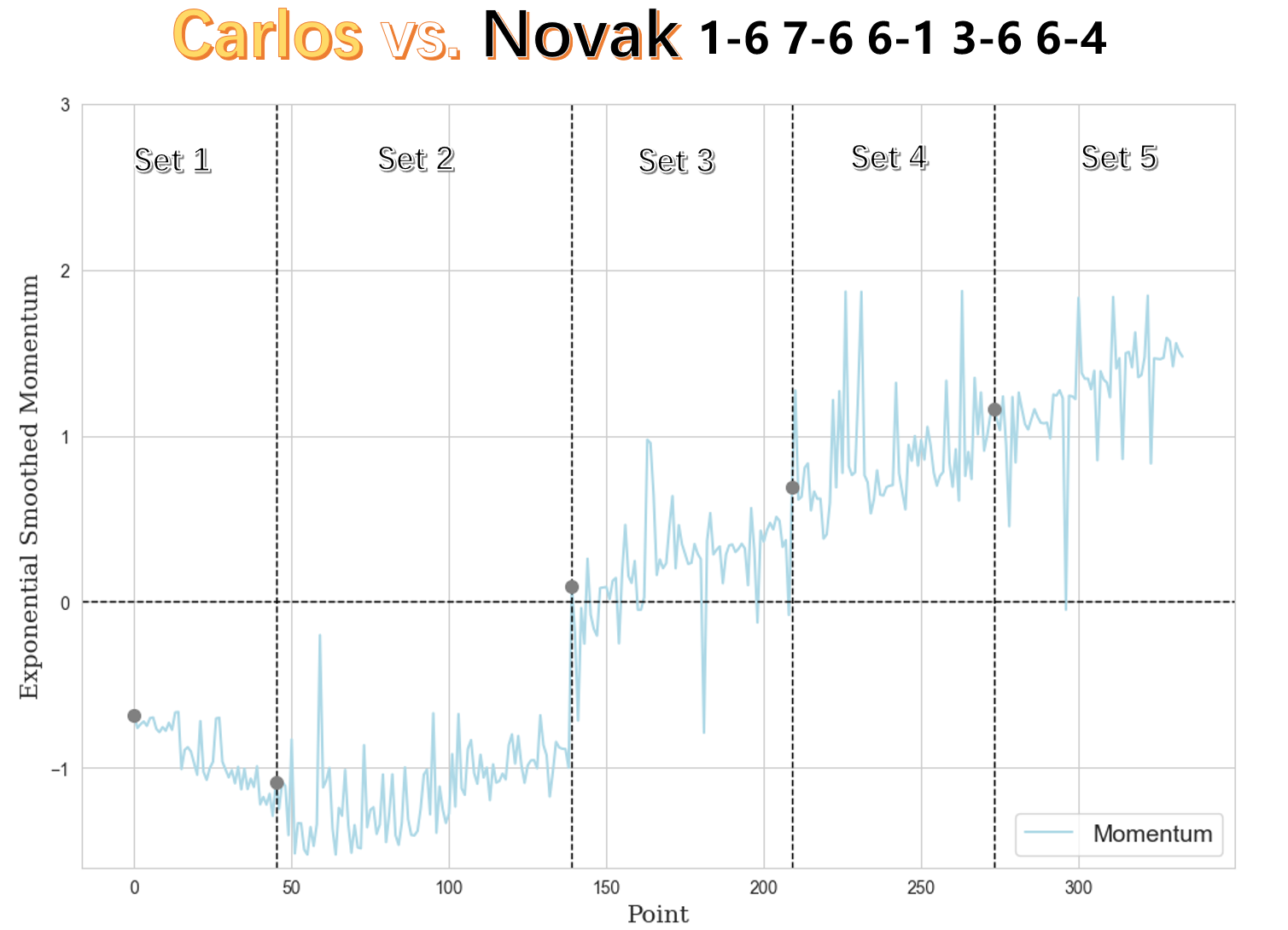}
			\caption{The Match Flow}\label{fig:1701}
		\end{subfigure}
	\caption{2023-Wimbledon-1701}
\end{figure}

First we use Pearson correlation coefficient to see what a player's probability of winning is related to. The result is in Figure \ref{fig:Pearson}. We see that momentum and probability of winning are closely correlated, suggesting that momentum exists and can be quantified; The high correlation between percentage that the player serves and probability of winning proves that the probability of winning is also somewhat related to the server.

After using HMM to obtain the state transition probability, we use EMA to obtain the quantitative value of momentum. We use the match data "2023-Wimbledon-1701" (the Wimbledon Gentlemen's final). The momentum when scoring during the match is as shown in Figure \ref{fig:1701}. The momentum of 0 on the vertical axis means that neither side has an advantage; [2,0) means that the momentum is in Djokovic, and [0,2) means that Alcarez has the momentum.

Figure \ref{fig:1701} depicts the match flow in detail.
\textbf{In set 1}, momentum was in Djokovic. Momentum smooth fluctuation was small, and showed an upward trend, so the score was 6-1, easy to take the first set. 
\textbf{In set 2}, both players kept scoring points and the game was back and forth. Djokovic's momentum was unstable, fluctuating greatly and showing a downward trend. In the end, he lost the set regrettably.
\textbf{In set 3}, the momentum favored Alcarez, and Alcarez's Momentum increased significantly. So he finally won the match by 6-1.
\textbf{In set 4}, although the overall momentum was in Alcarez, his momentum rise decreased and fluctuated more, but not as much as the momentum fluctuation when the score was 6-1, so he lost the match by 3-6.
\textbf{In set 5}, Alcarez had a higher momentum, greater than one. Because both players kept scoring, the momentum fluctuated. Finally Alcarez ended up winning the match 6-4.

\subsection{Performance Metrics}
We use the absolute size of momentum as players’ performance metrics.
Whichever athlete has momentum means that he performs better, and the greater the absolute value of momentum, the better the performance. Judging from the five sets, Alcarez's momentum was close to two at its highest, while Djokovic's momentum was only 1.6 at most. It shows that Djokovic performed better throughout the game.

\section{Assessing the role of Momentum} \label{b}
To assess the coach's claim, we use \textbf{Xgboost} Regression predictive Model. We construct one model where momentum is \textbf{Gaussian stochastic Process} and the other model where momentum is our proposed model in section \ref{model}. We predict momentum as a feature and build these two models for correlation analysis to assess the role of momentum. 
What's more, to compare these two models, we have established a series of \textbf{evaluation criteria}. And we also use \textbf{ROC curve} to intuitively express the difference between the two models.
\subsection{Xgboost Regression Model}
Xgboost is the abbreviation of "Extreme Gradient Boosting". The Xgboost algorithm is a type of synthetic algorithm that combines basis functions and weights to form a good data fitting effect.
Different from the traditional gradient boosting decision tree (GBDT), Xgboost adds a regularization term to the loss function, and since some loss functions are difficult to calculate the derivative, Xgboost uses the second-order Taylor expansion of the loss function as the fitting of the loss function.
Because Xgboost is more efficient when processing large-scale data sets and complex models, it also performs well in preventing overfitting and improving generalization capabilities. Therefore, since its introduction, Xgboost has been welcomed by the fields of statistics, data mining, and machine learning.

Xgboost is an algorithm based on gradient boosted trees. Gradient boosted trees are an ensemble learning method that iteratively trains a series of weak learners (usually decision trees) step by step. Each iteration attempts to correct the error of the previous iteration. Finally, these weak learners are combined into a strong learner.

For a dataset containing n entries of m dimensions, the Xgboost model can be represented as: 
\begin{equation}
	\begin{aligned}&
		\hat{y}_i=\sum_{k=1}^K f_k(x_i),\quad f_k\in F\quad(i=1,2,\ldots n)
\end{aligned}
\end{equation}

$F=\left\{f\left(x\right)=w_{q\left(x\right)}\right\}\left(q:R^m\to\left\{1,2,\ldots T\right\},w\in R^T\right)$ is a set of CART decision tree structures, q is the tree structure in which samples are mapped to leaf nodes, T is the number of leaf nodes, and w is the real score of leaf nodes.

\subsection{Momentum as Gaussian stochastic process}
\textbf{Gaussian stochastic process}
is an n-dimensional distribution of a random process that conforms to the normal distribution. A Gaussian process is statistically independent if it is uncorrelated at different times. The coach believes that "momentum" had no role and instead thinks that a player's fluctuations in play and success were random. So we build the predictive model considering momentum as a Gaussian stochastic process.

\subsection{Evaluation criteria and ROC curve} \label{criteria}
\begin{figure}[htbp]
	\centering
	\includegraphics[width=.3\textwidth]{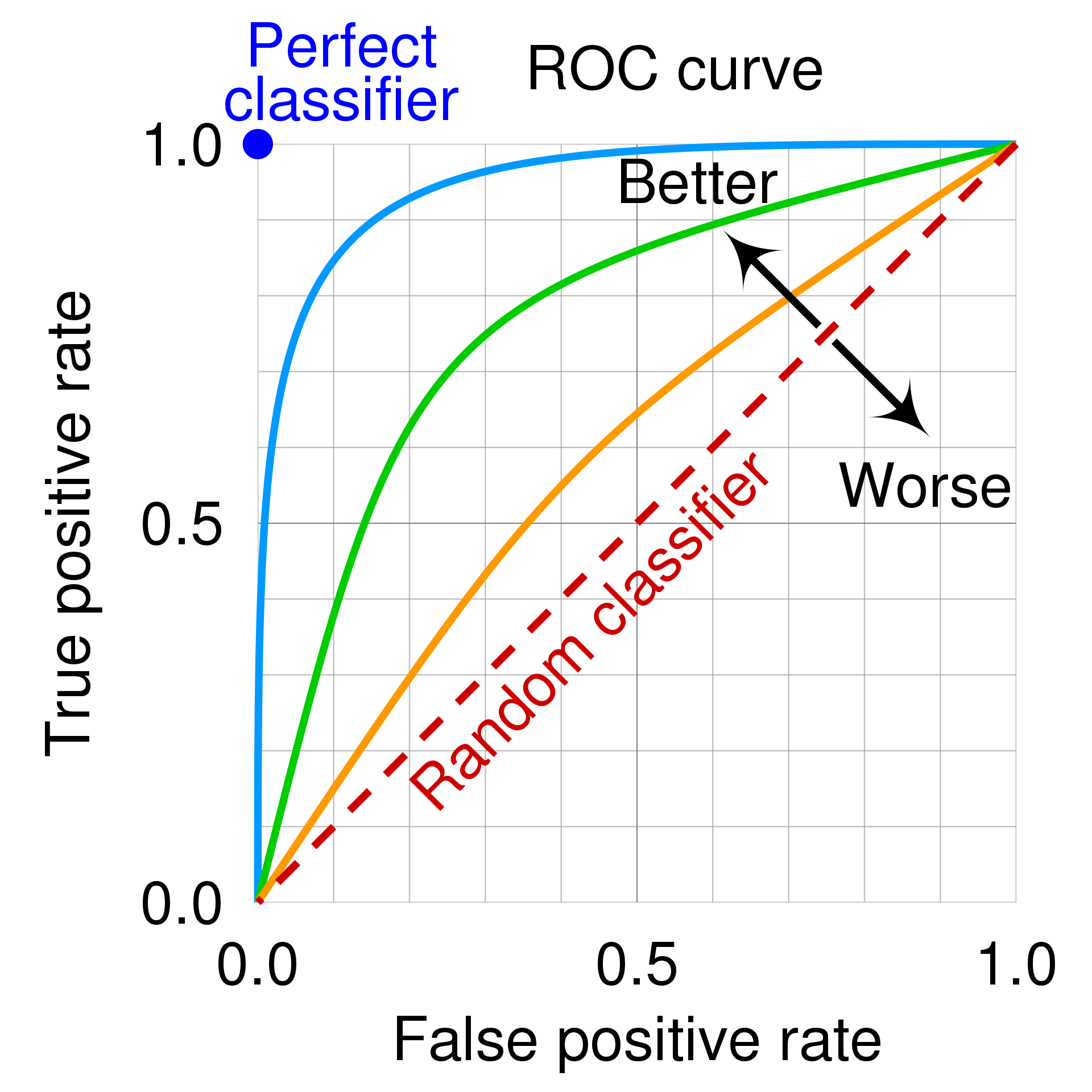}
	\caption{ROC curve's concept plot}\label{fig:ROC0}
\end{figure}

The ROC curve is a graphical tool for evaluating the performance of binary classification models. It plots a curve with the True Positive Rate (also known as recall or sensitivity) as the vertical axis and the False Positive Rate as the horizontal axis. The ROC curve demonstrates the performance of the model at different classification thresholds. The ideal shape of a ROC curve is a diagonal line from the lower left corner to the upper right corner, indicating good performance at all thresholds.
In general, the closer the curve is to the top of the diagonal, the better the model performance. The ROC curve's concept plot is shown in \ref{fig:ROC0}.

Confusion Matrix is an $N\times N$ table used to summarize the prediction effect of the classification model, where N represents the number of 
categories, the horizontal axis represents the actual labels, and the vertical axis represents the labels predicted by the model. A concept map is shown in Table \ref{tab:Confusion Matrix0}.
\begin{table}[htbp]
	\centering
	\caption{Confusion matrix concept map}
	\label{tab:Confusion Matrix0}
	\begin{tabular}{|c|c|c|c|l}
		\cline{1-4}
		\multicolumn{2}{|c|}{\multirow{2}{*}{}}        & \multicolumn{2}{c|}{\textbf{Actual}} &  \\ \cline{3-4}
		\multicolumn{2}{|c|}{}                         & Positive          & Negative         &  \\ \cline{1-4}
		\multirow{2}{*}{\textbf{Predicted}} & Positive & \textbf{TP}                & \textbf{FP}               &  \\ \cline{2-4}
		& Negative & \textbf{FN}                & \textbf{TN}               &  \\ \cline{1-4}
	\end{tabular}
\end{table}

TP = True Positive, TN = True Negative, FP = False Positive, FN = False Negative.

Based on confusion Matrix, we use the following indicators as model evaluation criteria.
\begin{enumerate}
	\item \textbf{Test Accuracy} is an indicator used to evaluate the performance of classification models. It is a measure of classification accuracy and represents the proportion of all samples that the model correctly classified. In binary classification problems, accuracy is calculated as: 
	$$\text{Accuracy}=\frac{TP+TN}{TP+TN+FF+FN}$$
	\item \textbf{AUC(Area Under the Curve)} is the area under the ROC curve, used to measure classifier performance from prediction scores. 
	$$AUC=P\left(P_\text{positive samples}{ > }P_\text{negative samples}\right)$$
	\begin{itemize}
		\item AUC = 1, which is a perfect classifier. When using this prediction model, there is at least one threshold that can produce perfect predictions. In most prediction situations, there is no perfect classifier.
		\item 0.5 < AUC < 1, better than random guessing. This classifier (model) can have predictive value if the threshold is properly set.
		\item AUC = 0.5, just like random guessing (eg: losing coins), the model has no predictive value.
		\item AUC < 0.5, worse than random guessing; but better than random guessing as long as it always goes against the prediction.
	\end{itemize}
	\item \textbf{Recall} represents the proportion of all positive examples in the sample that are correctly predicted. Used to evaluate the detection coverage of the detector for all targets to be detected. The recall calculation equation is shown in the following:
	$$Recall=R=\frac{\mathrm{\textit{TP}}}{ActualPositive}=\frac{TP}{TP+FN}$$
	\item \textbf{Precise} refers the precision rate is the frequency with which the model correctly predicts the positive class. Often, precision and recall are contradictory. The calculation equation is shown in the following:
	$$Precise=P=\frac{\mathrm{TP}}{PredictedPositive}=\frac{TP}{TP+FP}$$
	\item \textbf{F1 score:} Sometimes the difference between precision and recall is large, and the harmonic mean F1 score can reflect the overall algorithm performance. Usually precision, recall, and F1 score are used together to comprehensively reflect the performance of the algorithm. The equation is shown in the following:
	$$F\textit{1} ~ score=\frac{2}{\frac{1}{R}+\frac{1}{P}}=\frac{2*P*R}{P+R}=\frac{2TP}{2TP+FP+FN}$$
	
\end{enumerate}

\subsection{Comparisons and Conclusions}
\begin{figure}[htbp]
	\centering
	\begin{subfigure}[t]{.4\textwidth}
            \centering
		\includegraphics[width=\textwidth]{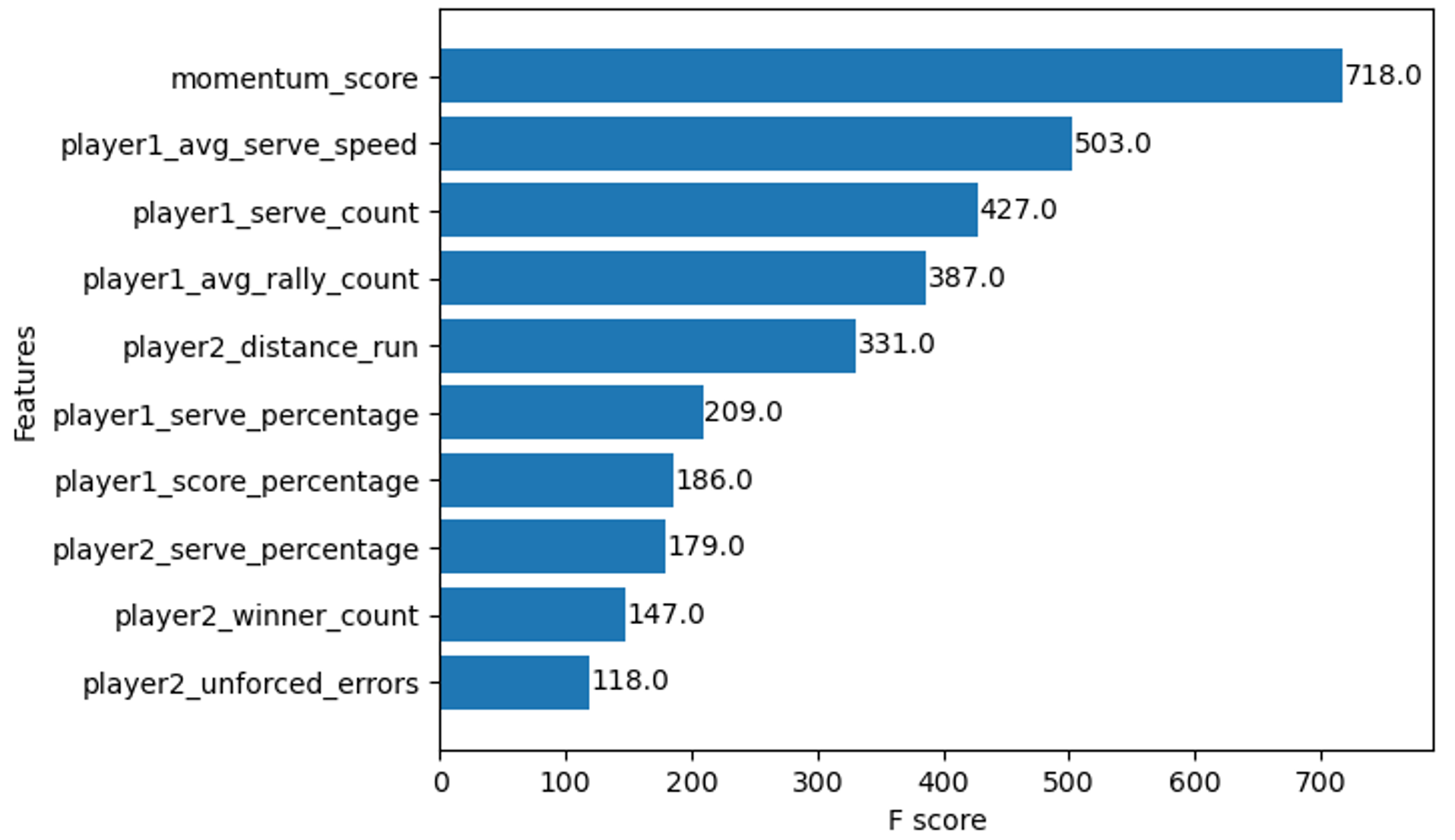}
		\caption{momentum is a random variable}
	\end{subfigure}
	\begin{subfigure}[t]{.4\textwidth}
            \centering
		\includegraphics[width=\textwidth]{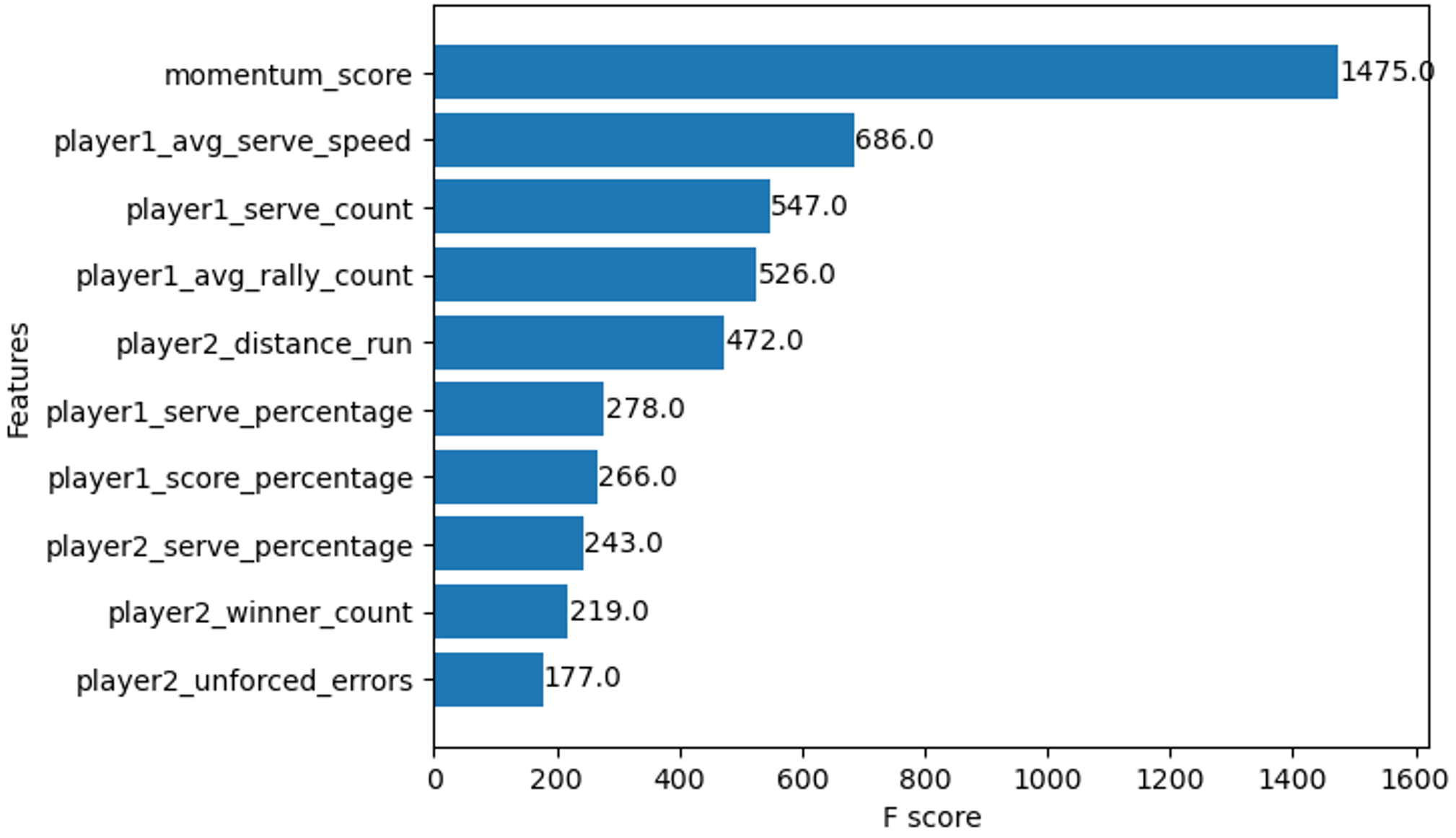}
		\caption{momentum is calculated by the model we proposed}
  \label{fig:111}
	\end{subfigure}
	\caption{feature importance from two models}\label{fig:XGB}
\end{figure}

Using Xgboost Regression, the feature importance diagrams of the two models are shown in Figure \ref{fig:XGB}. F score is the number of occurrences in the tree (as the number of dividing attributes). It can be seen that in Figure \ref{fig:111}, Momentum is much more important than other factors.

\begin{figure}[htbp]
	\centering
	\includegraphics[width=.4\textwidth]{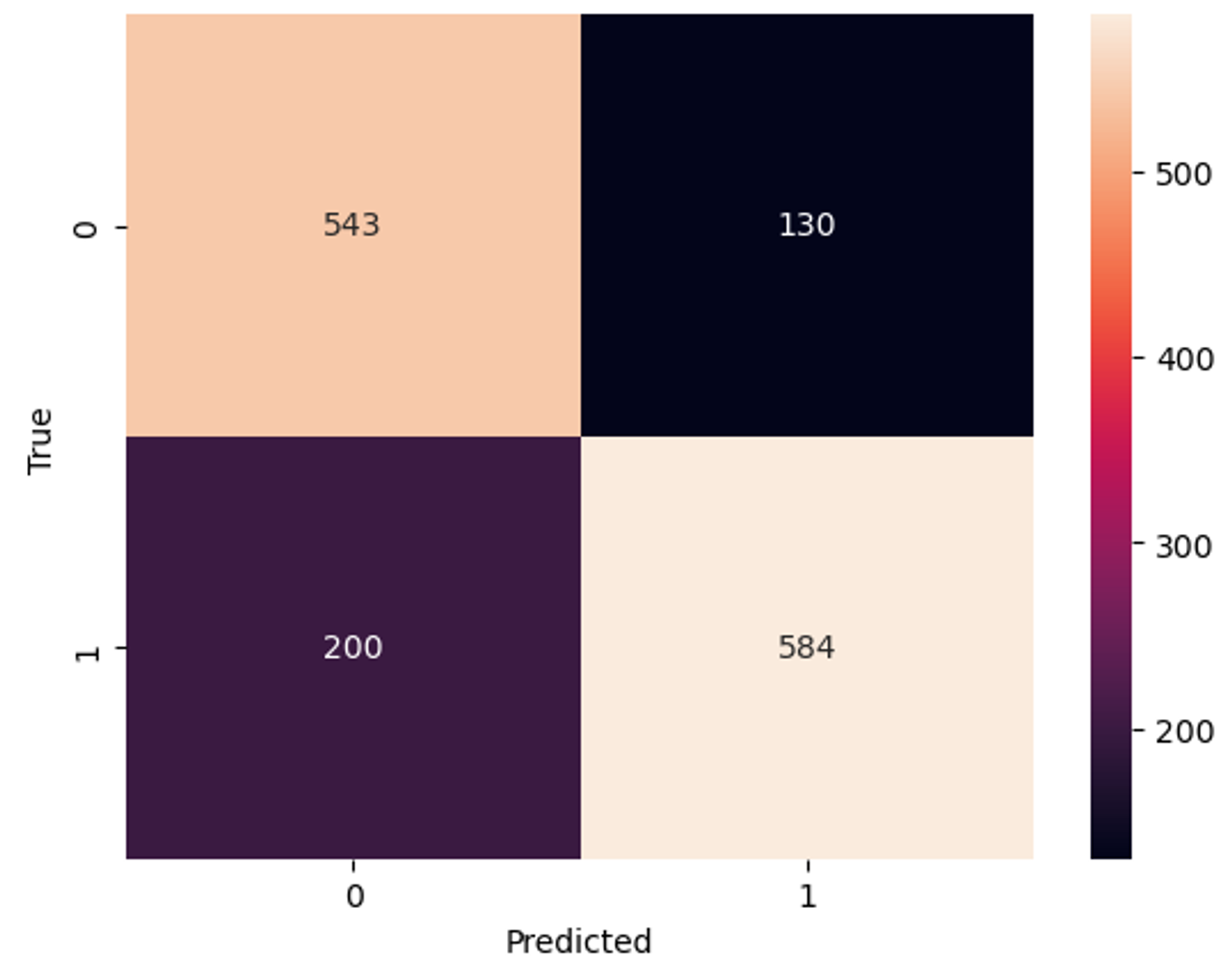}
	\caption{the confusion matrix}\label{fig:matrix}
\end{figure}

And the confusion matrix of the model considering momentum is shown in Figure \ref{fig:matrix}. 0 represents player 1, while 1 means player 2. It can be seen that TP and TN account for a large proportion, and the classification effect of momentum as a feature is better. This shows that under the prediction of this model, there is a high probability that the actual scoring player will be the same as the predicted player.

Comparing the two models, table 2 shows the model results when momentum is a Gaussian random variable, and table 3 shows the results when momentum is obtained from section \ref{model}.

\floatsetup[table]{capposition=top}
\newfloatcommand{capbtabbox}{table}[][\FBwidth]

\begin{table*}
\begin{floatrow}
\capbtabbox{
\begin{tabular}{ccc}
\toprule
Indicator & Value \\
\midrule
Test accuracy & 54.29\% \\
AUC & 54.28\% &  \\
recall & 52.98\% &  \\
F1 score & 53.43\% &  \\
precision & 53.88\% &  \\
\bottomrule
\end{tabular}
}{
\caption{}
 \label{Tab1}
}
\capbtabbox{
\begin{tabular}{ccc}
\toprule
 Indicator & Value \\
\midrule
Test accuracy & 81.30\% \\
AUC & 81.27\% &  \\
recall & 82.83\% &  \\
F1 score & 81.90\% &  \\
precision & 81.05\% &  \\
\bottomrule
\end{tabular}
}{
 \caption{}
 \label{Tab2}
 \small
}
\end{floatrow}
\end{table*}

It can be seen that when momentum is the model we proposed, the values of the evaluation indicators are higher than when momentum is a random variable. 0.5<auc<1 indicates that the model has better prediction ability. This also shows that momentum does have a role in the game. 

\begin{figure}[htbp]
	\centering
	\begin{subfigure}[t]{.35\textwidth}
			\includegraphics[width=\textwidth]{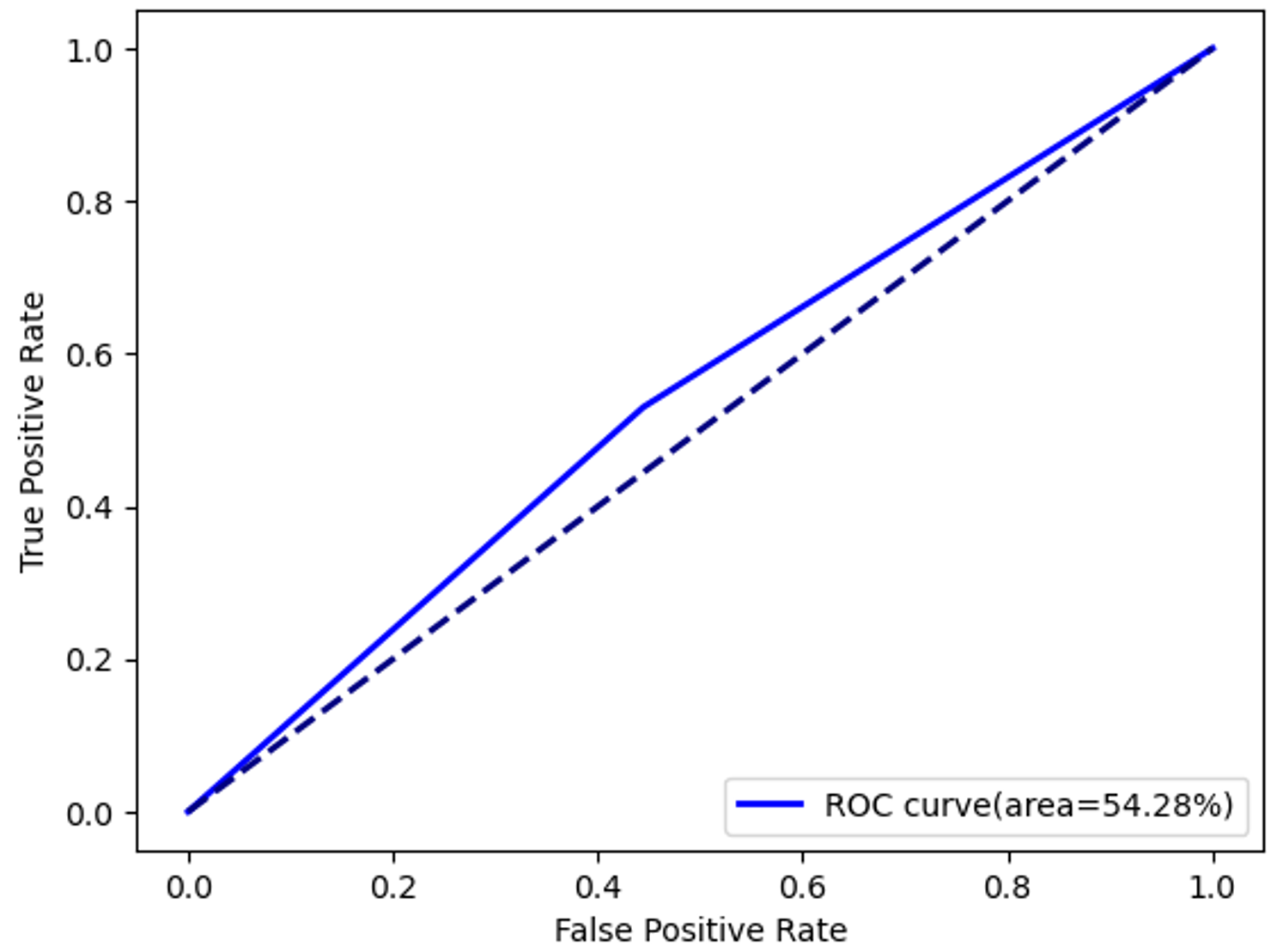}
			\caption{momentum is a random variable}\label{subfig:left}
		\end{subfigure}
	\begin{subfigure}[t]{.364\textwidth}
			\includegraphics[width=\textwidth]{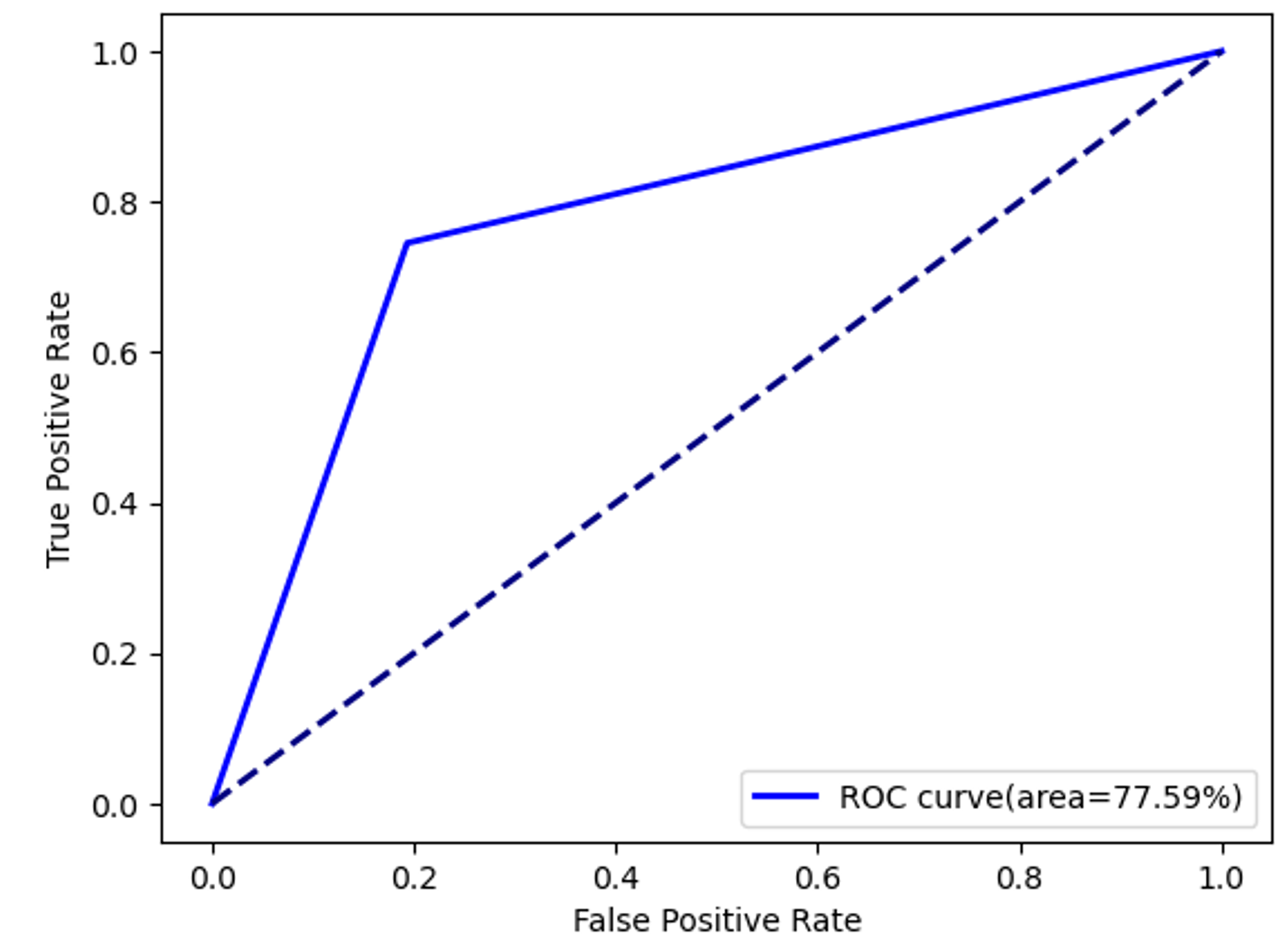}
			\caption{momentum is the model we proposed}\label{subfig:right}
		\end{subfigure}
	\caption{ROC curves}\label{fig:ROC}
\end{figure}

Similarly, we can compare ROC curves, obtained from the two models are shown in Figure \ref{fig:ROC}. From the determination rule of the ROC curve (Figure \ref{fig:ROC0}), we know that the model that takes into account the role of momentum has better prediction performance. We analyze the predictive power of the two models from a variety of perspectives and indicators. The results suggest that the coach's claim was inappropriate.

\section{Prediction Model and Indicator Selection}
\subsection{LightGBM Prediction Model}
\textbf{LightGBM} (Light Gradient Boosting Machine) is a framework that implements the \textbf{GBDT} algorithm and is a better version of Xgboost. It supports efficient parallel training and has the advantages of faster training speed and better accuracy.

\begin{figure}[htbp]
	\centering
	\includegraphics[width=.9\textwidth]{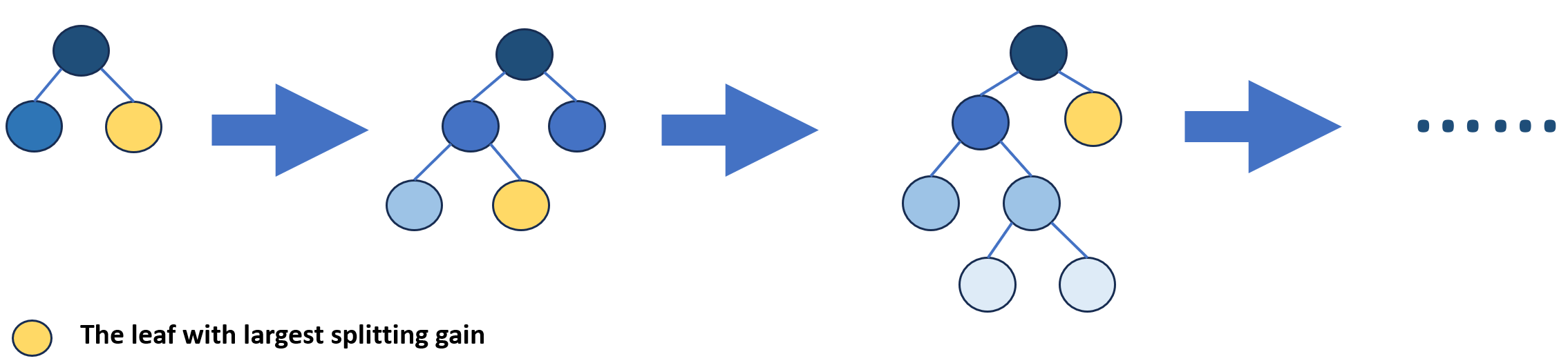}
	\caption{Decision Tree by Leaf-wise Growth}\label{fig:Leaf}
\end{figure}

On top of the Histogram algorithm, LightGBM is further optimized by abandoning the level-wise decision tree growth strategy used by most GBDT tools in favor of a leaf-wise algorithm with depth constraints. \textbf{Leaf-wise growth strategy} finds a leaf with the largest splitting gain each time from all the current leaves, then splits it, and so on. The process is shown in Figure \ref{fig:Leaf}. Therefore, compared with Level-wise, the advantages are: in the case of the same number of splits, Leaf-wise can reduce more errors and get better accuracy; the disadvantages are: it may grow a deeper decision tree and produce overfitting. Therefore, LightGBM adds a maximum depth limit on top of Leaf-wise to prevent overfitting while ensuring high efficiency.

Xgboost has 3 built-in common ways of calculating feature importance, namely "weight", "gain", and "cover". The meanings are shown below:
\begin{itemize}
	\item \textbf{weight:} weight form, indicating how many times a feature is used when splitting nodes in all trees.
	\item \textbf{gain:} average gain, indicating the average gain brought by a feature as a split node in all trees.
	\item \textbf{cover:} average coverage, indicating the average number of samples covered when a feature exists as a split node in all trees.
\end{itemize}

Comparing the three calculation methods, weight will give a higher value to the numerical feature because it has more variables and space to cut when the tree is split. Therefore, this indicator will mask important enumeration features. And we can also predict that in the same dimension, features with larger upper and lower bounds are more likely to be ranked top.

\begin{figure}[htbp]
	\centering
	\begin{subfigure}[t]{.4\textwidth}
			\includegraphics[width=\textwidth]{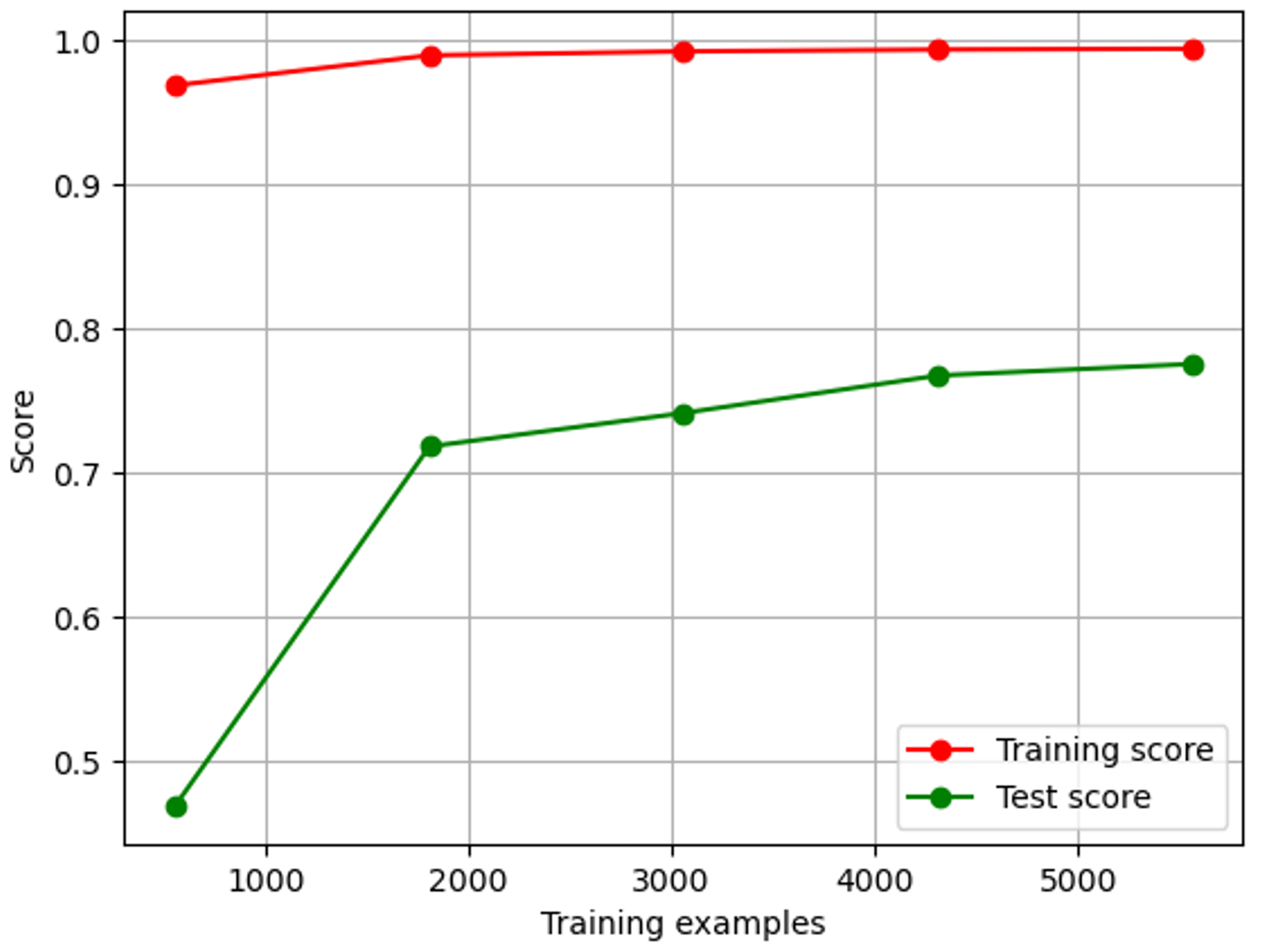}
			\caption{Model training process}\label{subfig:train}
		\end{subfigure}
	\begin{subfigure}[t]{.41\textwidth}
			\includegraphics[width=\textwidth]{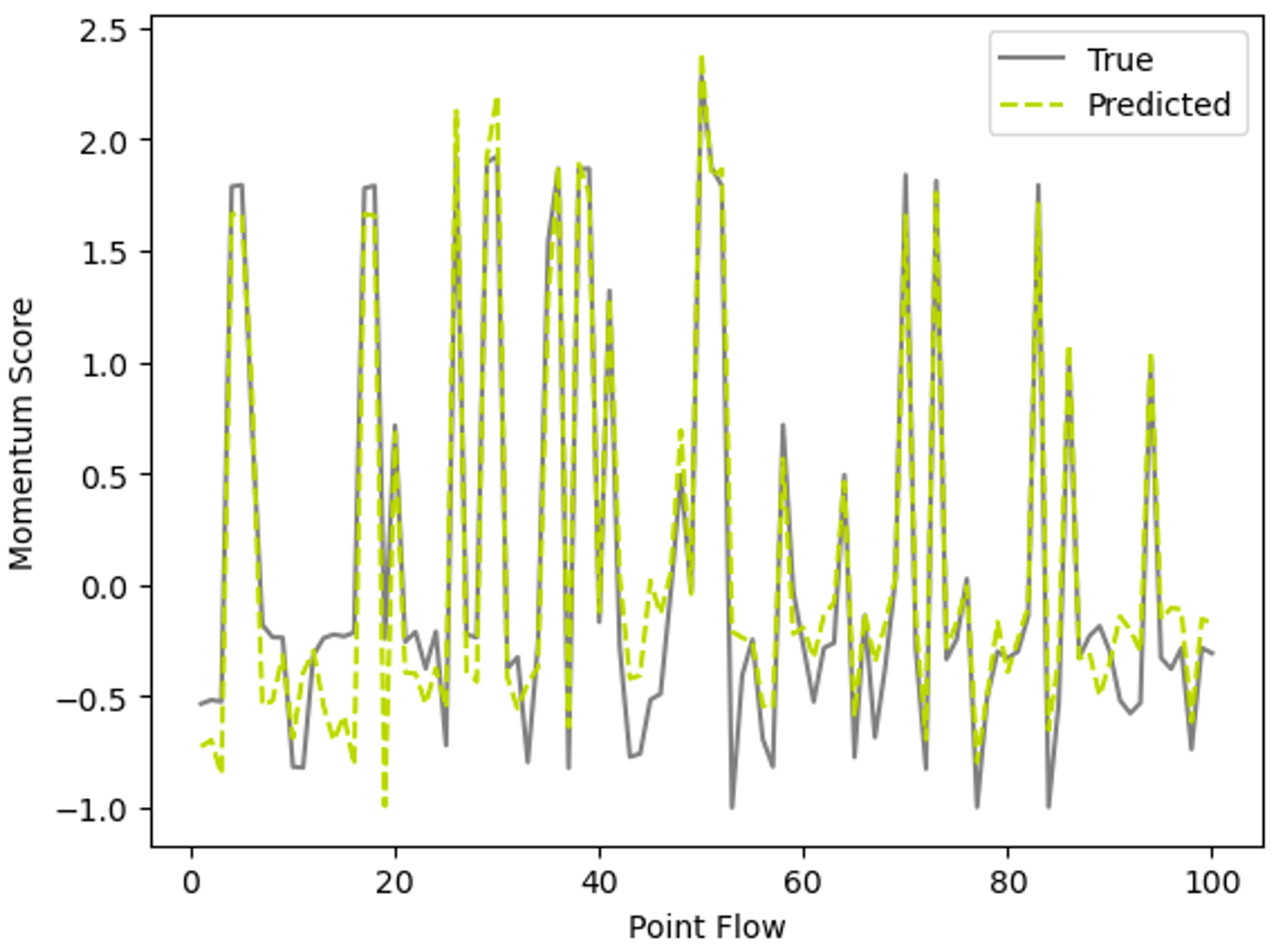}
			\caption{Momentum swings prediction result}\label{subfig:predict}
		\end{subfigure}
	\caption{LightGBM prediction model}\label{fig:sub}
\end{figure}

Based on 2023 Wimbledon Gentlemen’s final data, we use the LightGBM prediction model to predict momentum swings. The training process is shown in Figure \ref{subfig:train}. In order to make the picture clearer, we only plot one-third of the data prediction results of the match , as shown in Figure \ref{subfig:predict}.
It can be seen that the model test scores are very high. With a large sample size, the score is greater than 0.7. The accuracy of predicting momentum swings is also very good, with a high degree of overlap with the actual momentum change points.

\subsection{SHAP Method}
Since a big problem in machine learning is poor interpretability, although in \textbf{LightGBM} algorithm, feature\_importance can show the most important N features of the model, the situation for a single sample may not be consistent with the overall model. So we need methods for interpreting machine learning models and numerically displaying the contribution of different features in each sample. 
The methods for explanation are various, including LIME and SHAP. \textbf{LIME} is a local interpretation method, while SHAP is a global interpretation method. We ultimately chose to use the \textbf{SHAP} method to give feature importance.

SHapley Additive exPlanations (SHAP) is a technique to understand feature importance. The main idea behind the SHAP value is the Shapley value, a method from \textbf{the cooperative game theory}. Created by Shapley in 1953, the Shapley value is a method of allocating spending to players based on their contribution to the total spending. Players cooperate in a coalition and receive some benefit from this cooperation. To interpret machine learning predictions in terms of Shapley values, "total spending" is the model prediction for a single instance of the dataset, "player" is the instance's eigenvalue, and "gain" is the actual prediction of the instance minus the average prediction of all instances.

SHAP interprets the model's predicted value as the sum of the attributed values for each input feature \cite{10}:
$g(x^{\prime})=\phi_0+\sum_{j=1}^M\phi_jx_j^{\prime}$
, where $g$ is the explanatory model, $x_j^{\prime}$ indicates the corresponding feature, M is the number of input features, $\phi_j\in\mathbb{R}$ is the attribution value (Shapley value) of each feature and the calculation is equation \ref{phi}. $\phi_0$ is a constant in the model, and its value is equal to the predicted mean of all training samples.
\begin{equation} \label{phi}
	\phi_j=\sum_{S\subseteq\{x_1,\cdots,x_p\}\setminus\{x_j\}}\frac{|S|!(p-|S|-1)!}{p!}\left(f_x\left(S\cup\{x_j\}\right)-f_x(S)\right)
\end{equation}

$\{x_{1},\cdots,x_{p}\}$ is the set of all input features,
$p$ is the number of all input features,
$\{x_{1},\cdots,x_{p}\}\setminus\{x_{j}\}$ is the possible set of all input features excluding $\{x_{j}\}$,
$f_x(S)$ is the prediction of the feature subset $S$. It can be seen that the formula is the same as the definition of the Shapley value introduced earlier.

The specific steps of SHAP are as follows:
\begin{enumerate}
    \item Randomly extract the original data set to obtain multiple samples.
    \item For each sample, a complex model is used to predict and the prediction error is calculated.
    \item Based on the prediction error, select an optimal split point and divide the data set into multiple regions.
    \item Repeat steps 2-3 until the stopping condition is met.
\end{enumerate}

\subsection{Feature Importance}
\begin{figure}[htbp]
	\centering
	\begin{subfigure}[t]{.49\textwidth}
			\includegraphics[width=\textwidth]{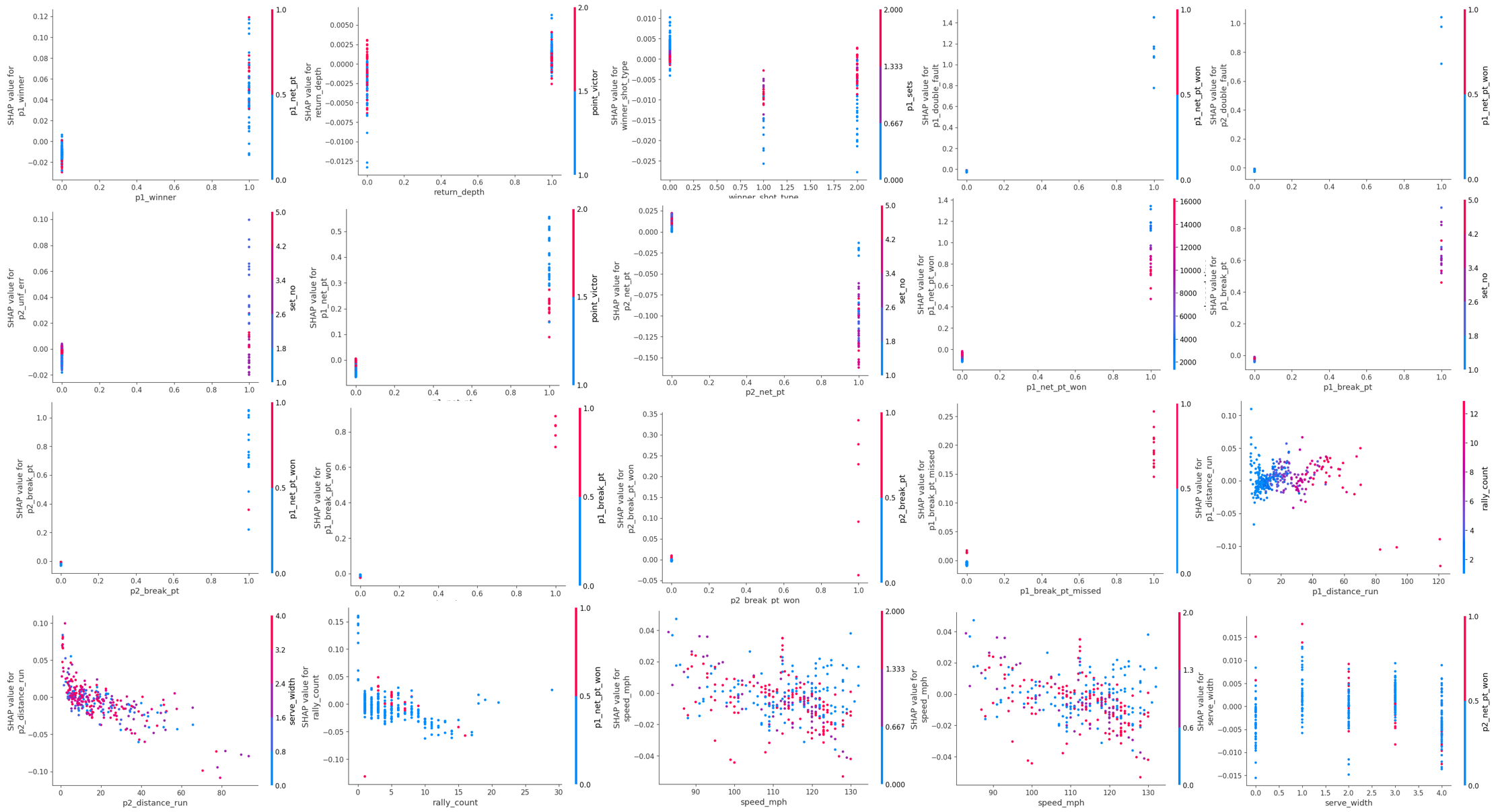}
		\end{subfigure}
	\begin{subfigure}[t]{.49\textwidth}
			\includegraphics[width=\textwidth]{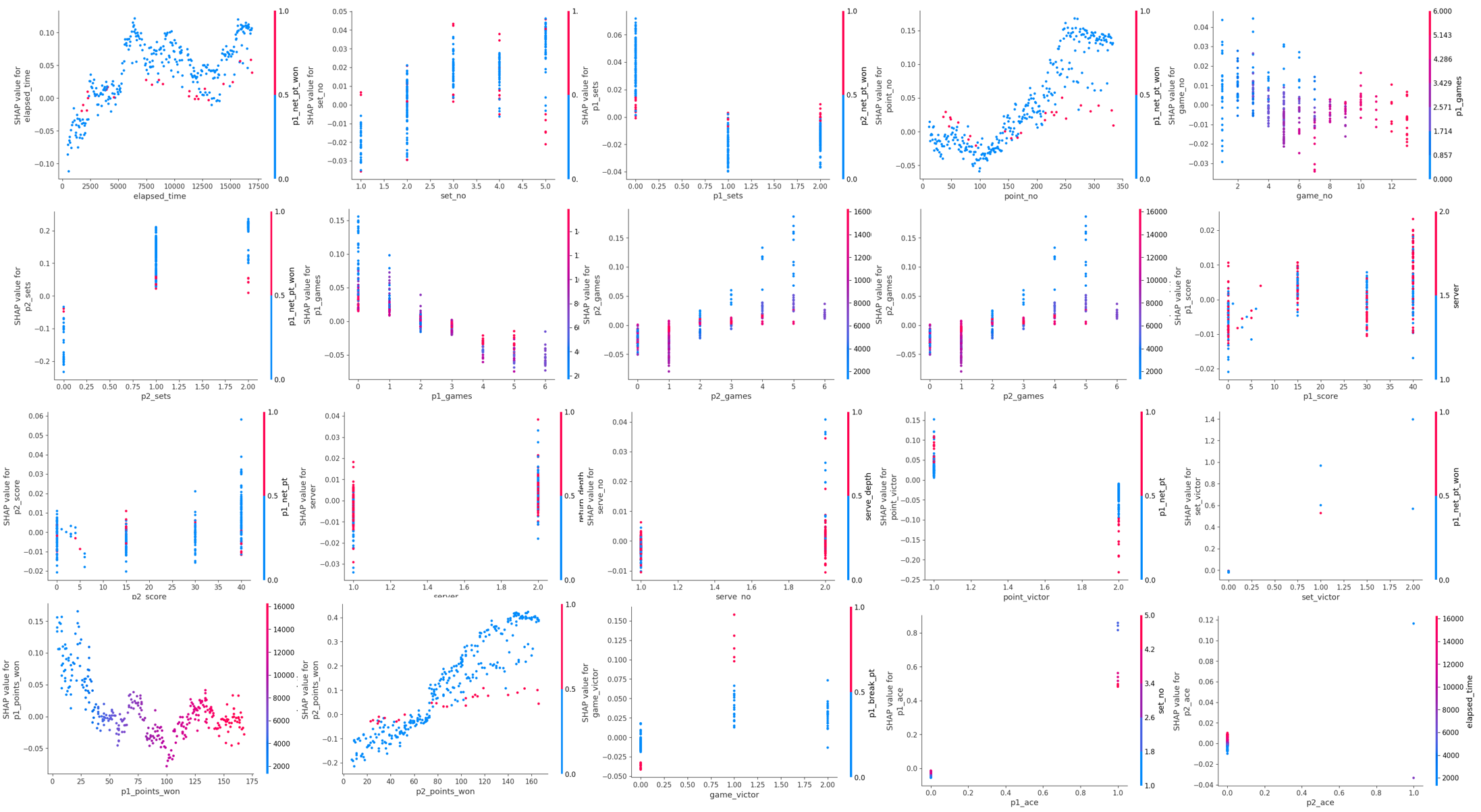}
		\end{subfigure}
	\caption{The joint action on Momentum of two different features}\label{fig:dot}
\end{figure}

First, we analyze the dependence between two features, that is, their joint contribution to momentum, and the impact of each two variables on the results. Figure \ref{fig:dot} shows the dependence.

\begin{figure}[htbp]
	\centering
	\begin{subfigure}[b]{.41\textwidth}
			\includegraphics[width=\textwidth]{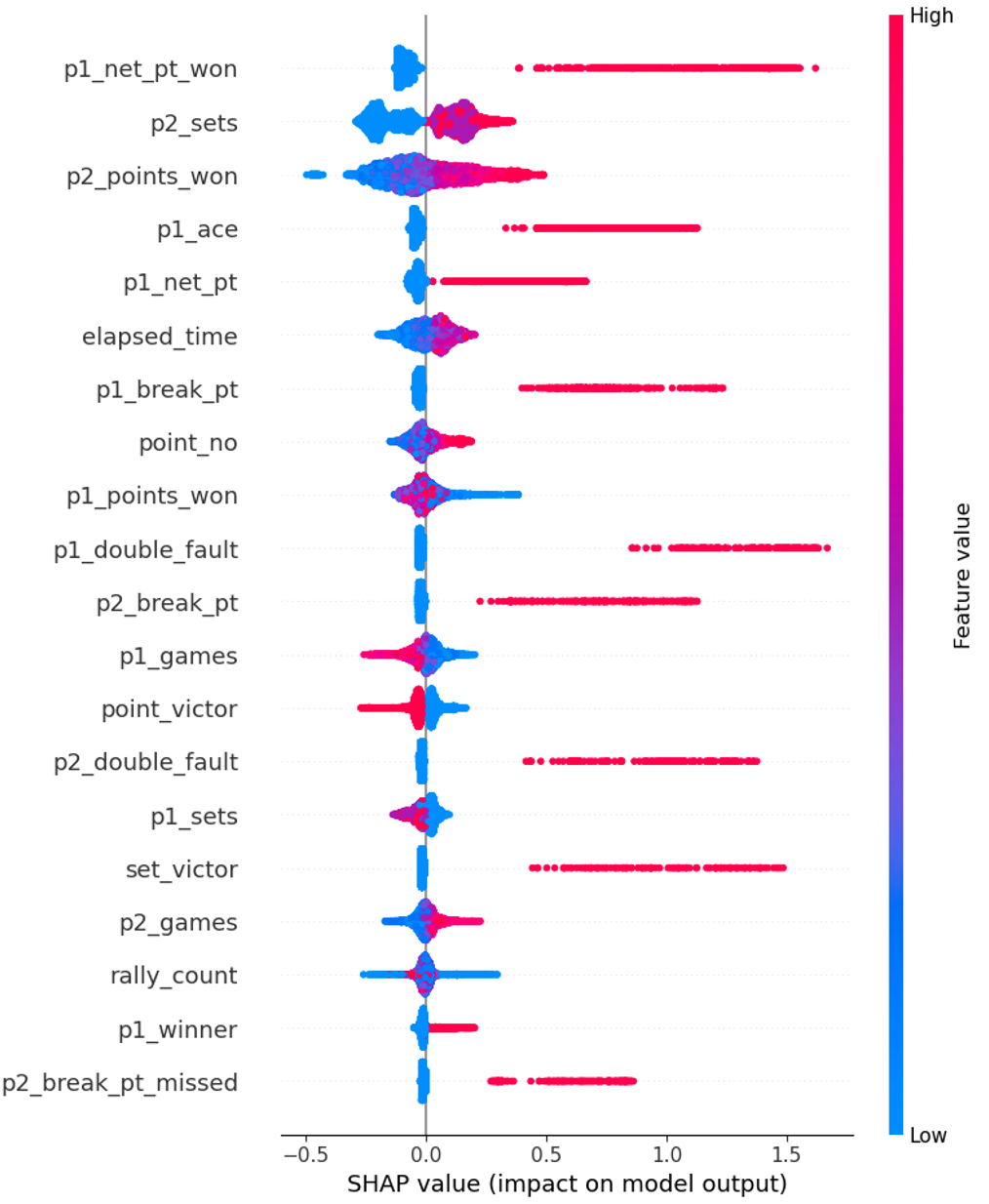}
			\caption{The SHAP value}\label{subfig:shap}
		\end{subfigure}
	\begin{subfigure}[b]{.4\textwidth}
			\includegraphics[width=\textwidth]{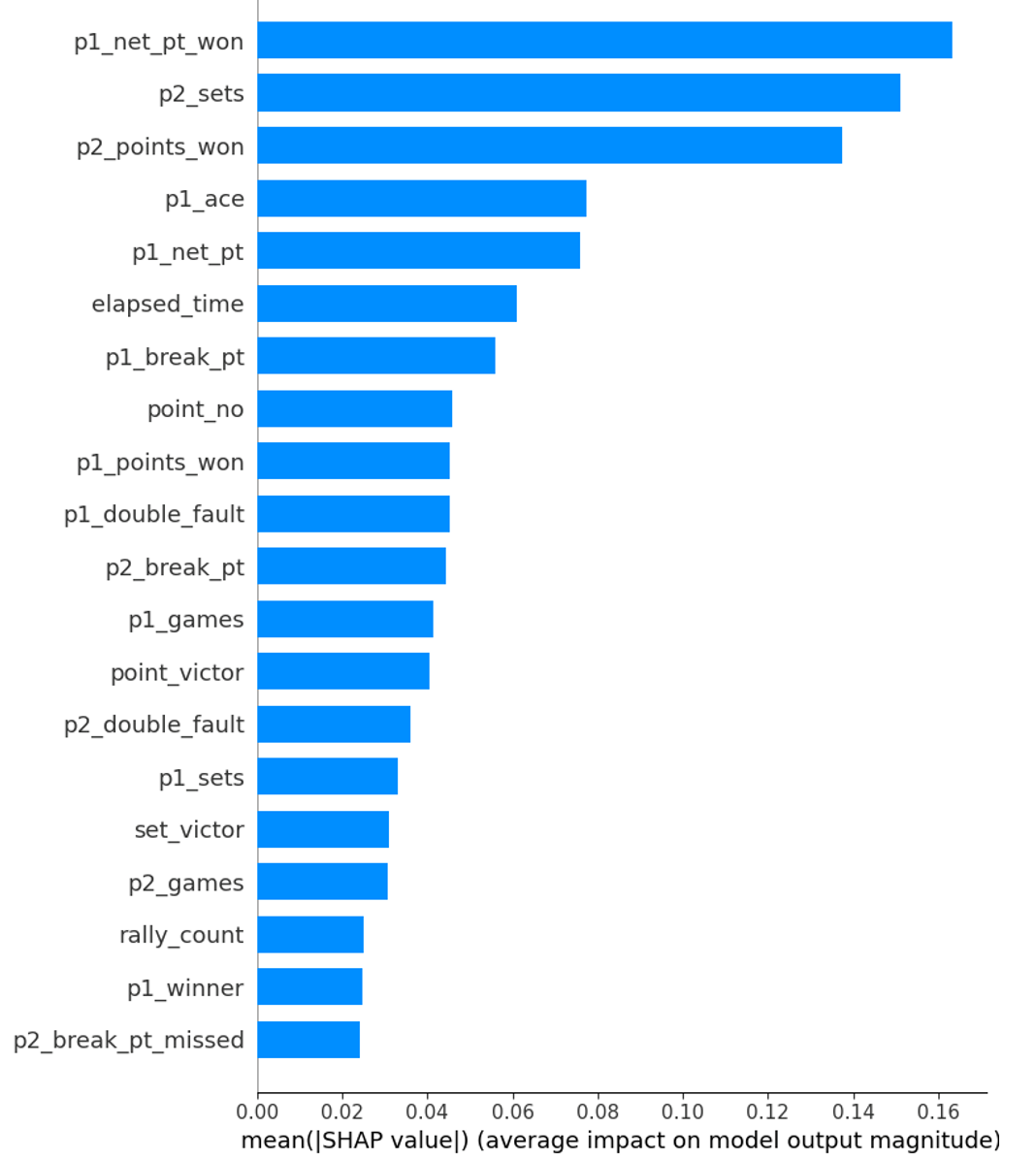}
			\caption{The importance ranking}\label{subfig:ranking}
		\end{subfigure}
	\caption{SHAP results}\label{SHAP results}
\end{figure}

Based on LightGBM and SHAP, we get SHAP value shown in Figure \ref{subfig:shap} and obtain the importance ranking of features as shown in Figure \ref{subfig:ranking}. 

In Figure \ref{subfig:shap}, the color of the dot represents the value of this feature for a certain sample (red is high, blue is low). The position of the dot represents the SHAP Value of the feature for a certain sample (left side is negative and right side is positive). A negative SHAP Value indicates that the feature is inversely related to the predicted value for this sample. In other words, for this sample, the larger the feature value, the smaller the predicted value. When the feature value is red, if the SHAP Value is greater than 0 (to the right of the central vertical line), the feature is generally more important.

Combining the results obtained, we choose the top-ranked features "p1\_net\_pt\_won", "p2\_sets", "p2\_points\_won", "p1\_ace", "p1\_net\_pt", "elapsed time", "p1\_break\_pt". Therefore, relying on these features as indicators can help determine when match points will occur and the timing of momentum changes.

\subsection{Suggestions for a new match}

    Considering the differences in momentum fluctuations in past matches, players can adopt different strategies when playing against different players in new matches. We will give Djokovic some advice since he lost the match we analyze.
\begin{itemize}
	\item Whenever you are behind or leading position, it is recommended that you should maintain a more stable state and pay attention to controlling the pace of the game to avoid being affected by yout opponents, you can also do something that will instill greater confidence in you since your points have won is a important feature in Changing the momentum.
	 \item When opponent tends to have some unforced errors, this suggests that he might not be fully focused at the moment, and now is the opportune time to catch him off guard, shifting momentum in your favor. Consistent training to enhance both physical endurance and hitting speed is crucial to capitalize on such opportunities during matches, since elapsed time and opponent's ace ball influence the momentum a lot.
	\item The ability to serve also seems to significantly impact your momentum. Therefore, it is crucial to enhance your practice in  serving regularly. Introducing artificial intelligence for higher-intensity training sessions can be advantageous. During competitions, preparation should rely on your judgment from experience and a comprehensive study of your opponent's habits, aiming to lead your opponent into your rhythm.
	\item When the opponent scores consecutively, the momentum swings greatly, or several parties make mistakes consecutively, the coach can call a timeout in time to discuss the tactics with the players.

\end{itemize}

\section{Model Evaluation}\label{d}
\subsection{Accuracy test}
\begin{figure}[htbp]
	\centering
	\includegraphics[width=.8\textwidth]{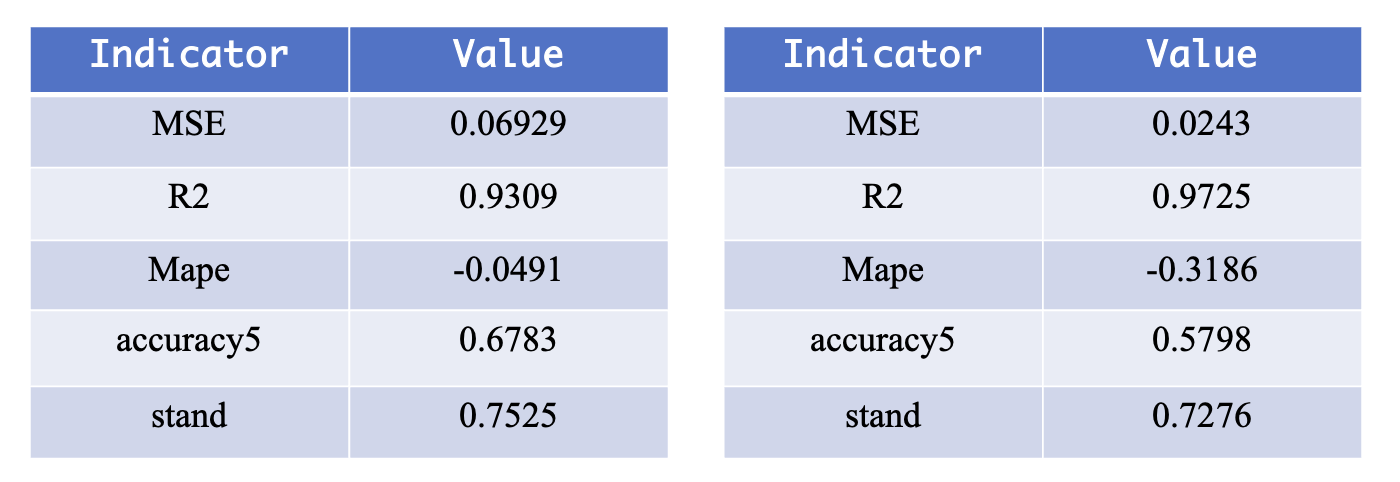}
	\caption{The prediction accuracy calculation}\label{fig:2}
\end{figure}

We selected two games for model testing, including the US Open data set 2023-usopen-1128 and 2023-usopen-1126.
In order to apply the model we built previously to the new competition dataset, we process the characteristics of this dataset to make it similar to the data we used previously. Then perform model testing. We choose a set of criteria for analysis.
The prediction accuracy is shown in Figure \ref{fig:2}. The left table is from 2023-usopen-1128, while the right table is 2023-usopen-1126. It can be seen that the prediction accuracy of our model in these two competitions is relatively high, indicating that our model has relatively good generalization.

\begin{figure}[htbp]
	\centering
	\begin{subfigure}[b]{.41\textwidth}
			\includegraphics[width=\textwidth]{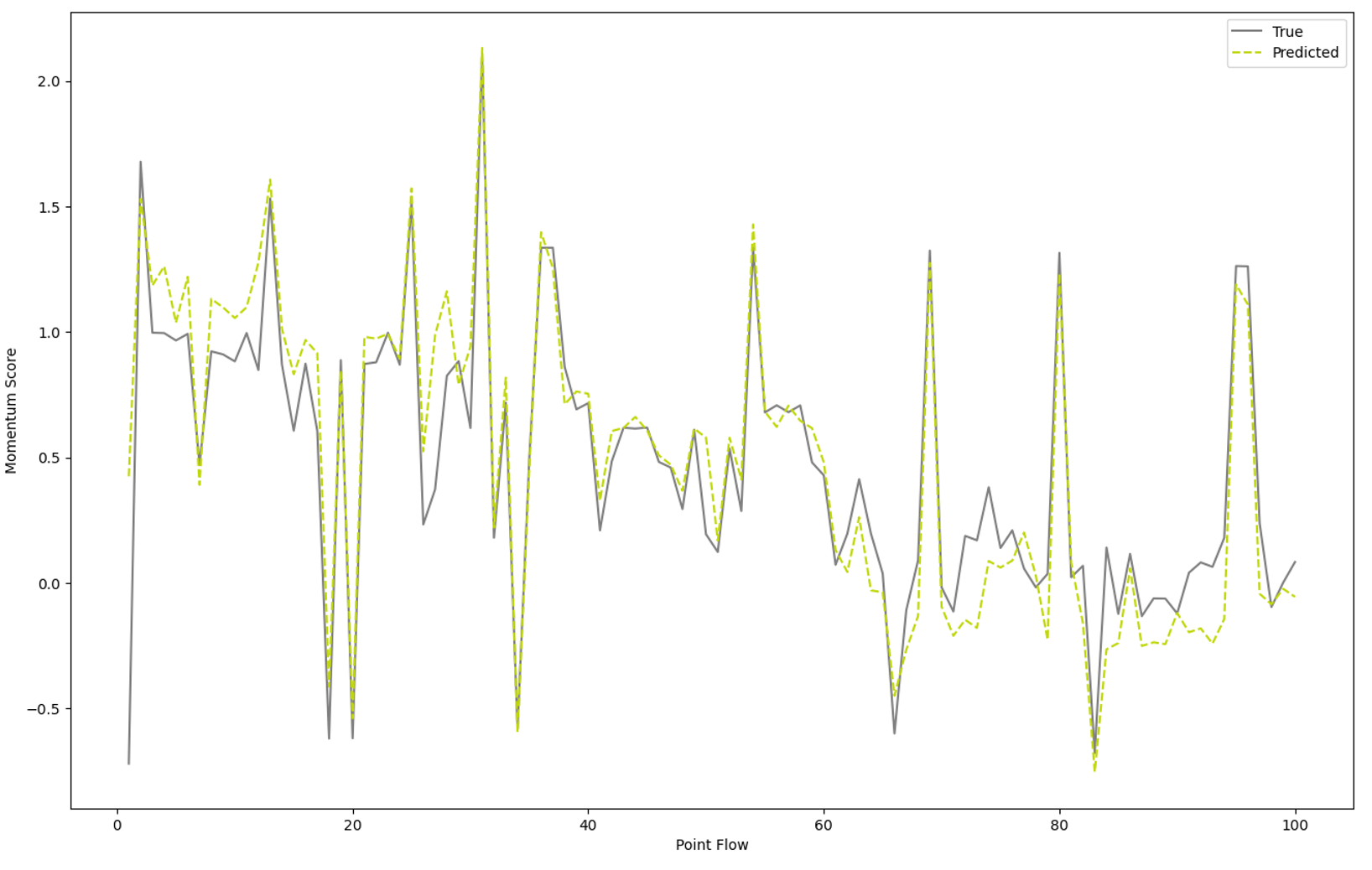}
			\caption{momentum prediction of 2023-usopen-1128}\label{subfig:b}
		\end{subfigure}
	\begin{subfigure}[b]{.41\textwidth}
			\includegraphics[width=\textwidth]{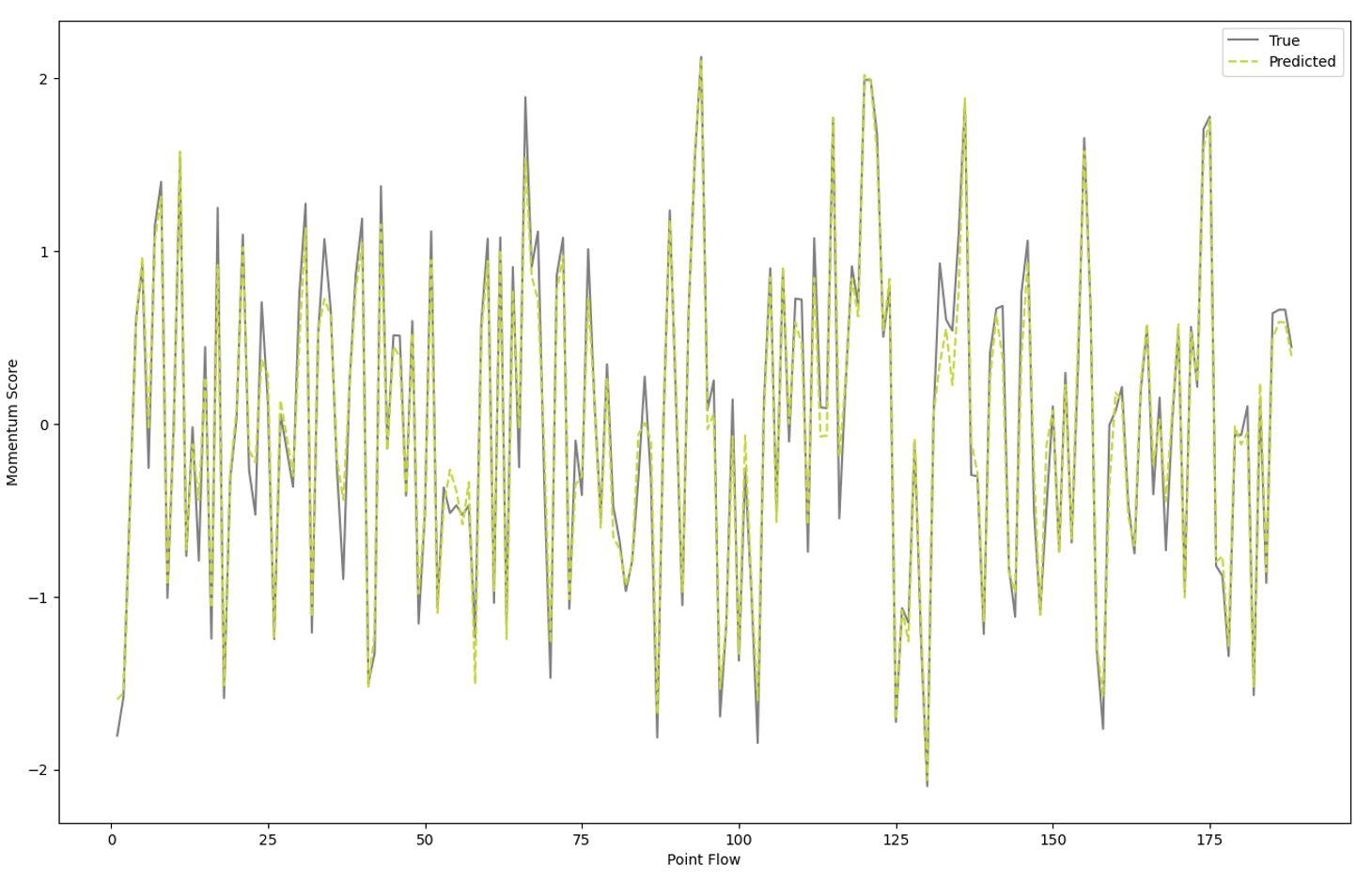}
			\caption{momentum prediction of 2023-usopen-1126}\label{subfig:c}
		\end{subfigure}
	\caption{Momentum swings prediction}\label{fig:a}
\end{figure}

The graph of predicting momentum change points is shown in the Figure \ref{fig:a}. The left figure is from 2023-usopen-1128 which is a men's match, while the right figure is 2023-usopen-2202 which is women's match. We can see that the predicted change value has a high degree of coincidence with the actual value, indicating that our model performs very well on these two data sets.

\subsection{Sobol Sensitivity Analysis}
We perform Sobol sensitivity method on lightGBM regression model used. 
We Use "SALib.analyze.sobol" to calculate the main effect and interaction effect of each independent variable and output the results. Among them, Si['S1'] is the main effect, indicating the contribution of each independent variable to the dependent variable; Si['ST'] is the total effect, indicating the contribution of all independent variables. S1: contains each input parameter The first type of Sobol index. ST: Contains the overall change index for each input parameter. The blue line is S1, the black line is ST. The result is the following Figure \ref{fig:sobol}. It can be concluded that our model has good robustness under specific variable combinations.
\begin{figure}[htbp]
	\centering
	\includegraphics[width=.8\textwidth]{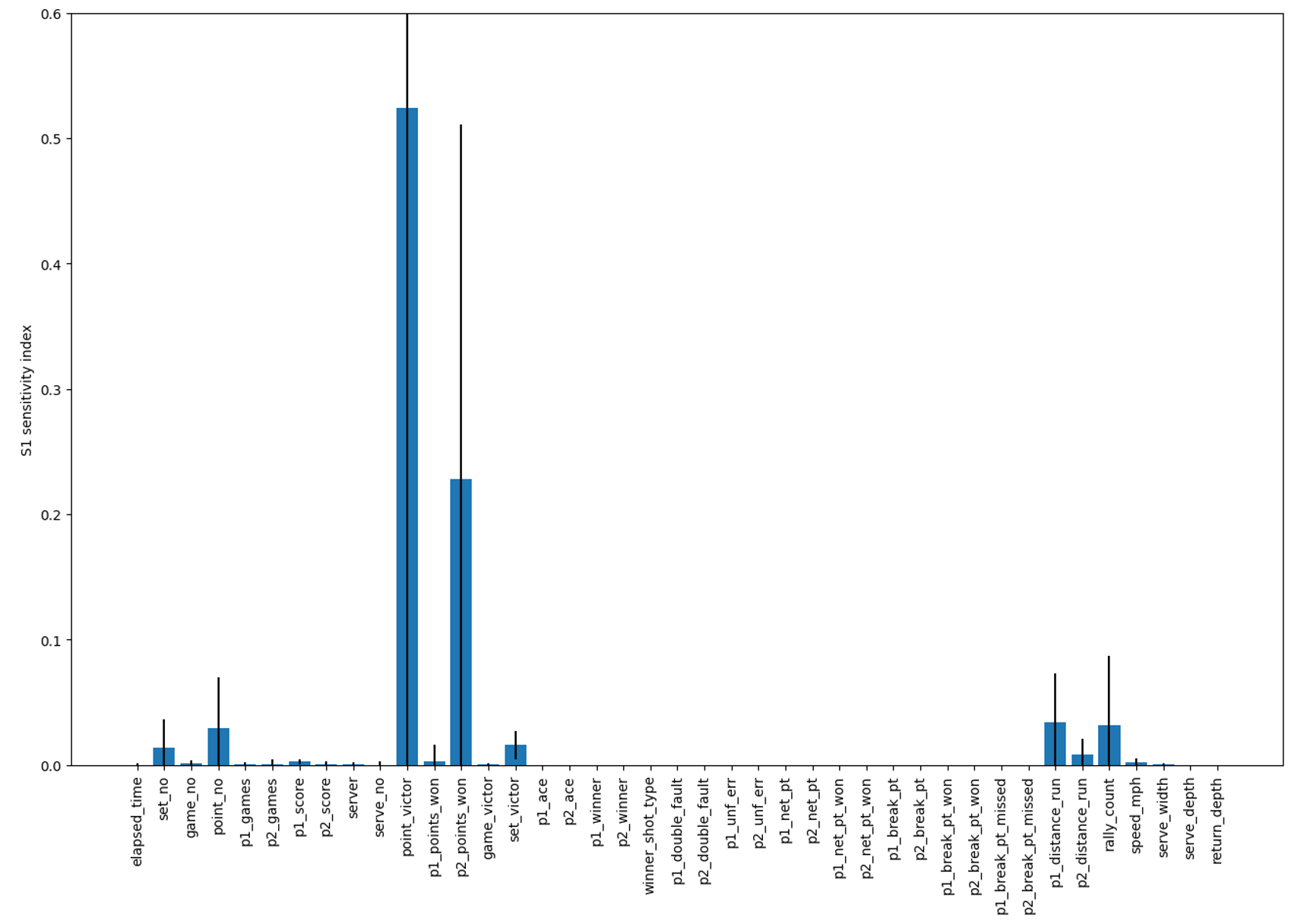}
	\caption{Sensitivity Analysis result}\label{fig:sobol}
\end{figure}

\subsection{Potential factors}
There are many factors that also affect the score and momentum of the match.

\textbf{Ranking:} Player world rankings, player world ranking points, etc. will affect the players' mental state and the match scores.

\textbf{Recent matches:} The player's recent match status can be obtained from the player's recent match wins, thus making the model prediction more accurate.

\textbf{Objective conditions:} The age of the player, the importance of the match, the material of the playing field and the betting odds of the player before the match, etc. will have an impact on the player's performance and the result of the match.

In improving the model in the future, we can consider the above factors to make the model predictions more accurate.

\section{Conclusion}
This memo discusses the problem of momentum and suggestions on preparation for match. Our team has developed a model on the momentum of tennis matches. It can predict accurately the match flow and the momentum swings. The model also can analyze the determining factors of the probability of a player winning the match. What' more, the model has a certain generalization ability, which means it can be applied to other matches. Based on our model and the results we get, we would like to give you some advice about tennis matches to help players better prepare for their matches and get good results.

\begin{itemize}
	\item \textbf{The role of momentum}
	\begin{enumerate}
	    \item Our analysis shows that momentum does play an important role in the game. An increase in momentum is associated with improved player performance, while a decrease in momentum may signal a shift in the flow of the game. You can use our model to \textbf{predict momentum} and adjust your game strategy in a timely manner.
	    \item Our model can \textbf{analyze the indicators of momentum swings}. 
	    It can be used in daily training to analyze the players and thus adjust the training content. What's more, it can also be used to analyze the opponent's serving and winning characteristics. This allows players to better prepare for matches.
	\end{enumerate}

	\item \textbf{Suggestions before the match} 
	
	The model we propose can be used to analyze the data of multiple matches of the opponent or one's own players. Then find out the factors that usually determine the player'winning in matches. Identify patterns or weaknesses that may trigger a swing in momentum, so that you can make targeted preparations before the match.
	\begin{enumerate}
	\item We find that the feature with the highest importance in momentum swings is the points won at the net. That means the player has a high probability of scoring when he touches the net. Therefore, the player should seize the opportunity to serve during the match, while his opponent needs to conduct training for the ball that touches the net. 
	\item By using our model, the indicators of each player'momentum swings can be analyzed. Thus carry out targeted training. Our results show that the current number of points won is ranked higher in importance. That shows the player is easily affected by the score. When he scores more points, the probability of winning will be higher. So the player needs to train for mental issues. 
	\item The higher ranked ones also have ace, which shows that players need to train more on hitting skills, develop new techniques, and improve their ace abilities during their regular training.
	\end{enumerate}
	\item \textbf{Suggestions during the match}
	\begin{enumerate}
		 
		\item Strengthen mental training, especially in the face of momentum swings, need to remain calm and focused. Our results show that one of the top ranked indicators on the flow of the match is "\textbf{elapsed time}". This shows that if the opponent is eager to find scoring points when the momentum swings, the players can calm down and look for the opponent's flaws without being too anxious to score.	You can also simulate momentum swings in training to improve the adaptability of players.
		\item Focus on specific \textbf{momentum-turning point} indicators, such as point differentials, serve breaks, etc., which may signal an impending change in the flow of the match. 
		Tennis break points are key scores that can affect the flow of the match and even the outcome of the match. In our results, break points have top rankings. This shows that when a player doesn't have the advantage of serving, he needs to seize the opportunity to utilize break points to make the situation turn around.
		\item Train players to recognize and utilize \textbf{momentum} changing points in the match, such as mental preparation and tactical adjustments when scoring consecutive points or facing important score. At the same time, the player should also pay attention to the opponent's status and look for scoring points.
	\end{enumerate}
\end{itemize}





%
%
%
%
%

\end{document}